\pdfoutput=1

\documentclass[11pt]{article}

\usepackage[preprint]{acl}

\usepackage{times}
\usepackage{latexsym}

\usepackage{booktabs}
\usepackage{multirow}
\usepackage{colortbl} 
\usepackage{tcolorbox}              
\usepackage{markdown}

\usepackage[T1]{fontenc}

\usepackage[utf8]{inputenc}

\usepackage{microtype}

\usepackage{inconsolata}

\usepackage{graphicx}

\usepackage{epigraph}
\usepackage{booktabs}
\usepackage{multirow}
\usepackage{tikz}
\usepackage{amssymb} 
\newtheorem{remark}{Remark}

\newcommand*\circled[1]{\tikz[baseline=(char.base)]{
            \node[shape=circle,draw,inner sep=0.25pt] (char) {#1};}}
\newcommand*\circledspace[1]{\circled{#1}\hspace{0.5em}}

\title{From Chaos to Order: The Atomic Reasoner Framework \\for Fine-grained Reasoning in Large Language Models}

\author{
  \textbf{Jinyi Liu\textsuperscript{1}},
  \textbf{Yan Zheng\textsuperscript{1,†}},
  \textbf{Rong Cheng\textsuperscript{1,†}},
  \textbf{Qiyu Wu\textsuperscript{1,†}},
  \textbf{Wei Guo\textsuperscript{1}},
  \textbf{Fei Ni\textsuperscript{1}},\\
  \textbf{Hebin Liang\textsuperscript{1}},
  \textbf{Yifu Yuan\textsuperscript{1}},
  \textbf{Hangyu Mao\textsuperscript{2}},
  \textbf{Fuzheng Zhang\textsuperscript{2}},
  \textbf{Jianye Hao\textsuperscript{1}}
\\
  \textsuperscript{1}Tianjin University,
  \textsuperscript{2}Independent Researcher
\\
  \small{
      \textsuperscript{†}Equal contribution
  }
}

\begin{document}
\maketitle
\begin{abstract}

Recent advances in large language models (LLMs) have shown remarkable progress, yet their capacity for logical ``slow-thinking'' reasoning persists as a critical research frontier. Current inference scaling paradigms suffer from two fundamental constraints: fragmented thought flows compromising logical coherence, and intensively computational complexity that escalates with search space dimensions. To overcome these limitations, we present \textbf{Atomic Reasoner} (\textbf{AR}), a cognitive inference strategy that enables fine-grained reasoning through systematic atomic-level operations. 
AR decomposes the reasoning process into atomic cognitive units, employing a cognitive routing mechanism to dynamically construct reasoning representations and orchestrate inference pathways. This systematic methodology implements stepwise, structured cognition, which ensures logical coherence while significantly reducing cognitive load, effectively simulating the cognitive patterns observed in human deep thinking processes.
Extensive experimental results demonstrate AR's superior reasoning capabilities without the computational burden of exhaustive solution searches, particularly excelling in linguistic logic puzzles. These findings substantiate AR's effectiveness in enhancing LLMs' capacity for robust, long-sequence logical reasoning and deliberation.

\end{abstract}

\section{Introduction}
\epigraph{Knowledge is limited, whereas imagination embraces the entire world, stimulating progress, giving birth to evolution.}{--- Albert Einstein}

Large language models (LLMs) have demonstrated impressive capabilities in long-term thinking and reasoning, to solve complex problems through extended, logical thought processes. Especially, the emergence of OpenAI's o1 series models~\cite{o1} has sparked significant interest among researchers in replicating such slow-thinking, deliberative reasoning abilities.

Recent research has explored two primary approaches to achieve this goal. The first approach emphasizes training scaling through techniques such as Reinforcement Fine-Tuning (RFT)~\cite{trung2024reft}, employing reinforcement learning ~(RL) algorithms~\cite{sutton2018reinforcement} to optimize the LLM parameters guided by meticulously designed reward signals. Models developed under this paradigm, like QwQ~\cite{qwq}, kimi-1.5~\cite{team2025kimi1.5}, and Deepseek-R1~\cite{guo2025deepseekr1}, have demonstrated remarkable generalization capabilities. However, training-based methods are heavily dependent on training resources.
In contrast, alternative research has shifted focus from training mechanism to test-time inference scaling. This involves designing sophisticated mechanisms to guide or search for generting systematic, step-by-step thinking~
\cite{yanplanning,jiang2024technical,o1journey}. 
Inference scaling is widely regarded as a promising approach to achieving faster enhancements in performance of LLMs~\cite{snell2024scaling}.

Nevertheless, recent progress has pointed out, the efficacy of inference does not fundamentally reside in the specific search framework, but rather in the quality of guidance signals (typically reward models) and the effort in reasoning computational cost~\cite{gan2025rethinking,chen2024tree}. The former necessitates task-specific reward model training for different domain, while the latter necessitates extensive chain-of-thought expansion and exploration across vast solution spaces, both of which pose significant challenges to reasoning efficiency. 
On the other hand, the reasoning process often lacks fine-grained directional guidance, and such an open-ended generative process tends to induce high stochasticity in logical coherence. Therefore, this study focuses on a critical question: \textit{Can we achieve more efficient O1-like slow thinking with controllable reasoning costs through finer-grained and more intelligent guidance?}

To achieve human-like slow-thinking reasoning and address the challenges posed by inference scaling, this study proposes a novel reasoning framework called \textbf{Atomic Reasoner~(AR)}. This framework atomizes the reasoning process and incorporates a cognitive routing mechanism to guide LLMs in a refined, systematic and structured manner for high-quality slow-thinking reasoning.

Considering that solving complex problems requires a hierarchical thinking structure, the separation of the higher-order guidance (which determines the direction of thinking) and the execution (which handles specific reasoning steps) is a necessary architecture.
Firstly, to offer detailed and necessary guidance and constrain the solution space, we atomize the reasoning process and define a set of atomic reasoning actions~(ARAs). Among them, to emphasize the importance of premise information, we distinguish it for focused attention. This aligns with the philosophical and cognitive science characterization of the human slow thinking process~\cite{kahneman2011thinking,dewey1997we}.
Then, to organize the complex reasoning process, we introduce the Atomic Tree, enhancing its transparency and clarity.

To provide precise guidance throughout the reasoning process, we design an efficient cognitive routing mechanism. This mechanism guides the LLM to select appropriate reasoning actions based on the reasoning process represented by the atomic tree, providing fine-grained guidance for problem-solving. This guidance can reduce the solution space that the LLM explores at that step, facilitating the generation of more informative reasoning outcomes. Also, the tree-based cognitive routing mechanism directs the LLM to continuously expand the reasoning pathways, backtrack, branch out, and ultimately conclude the reasoning process. This approach realizes an anthropomorphic, deliberate slow-thinking process.

Experimental results indicate that AR effectively stimulates the ability of LLM to engage in complex slow-thinking reasoning. On the other hand, the high-quality reasoning data generated by AR can effectively enhance the base LLM's complex logical reasoning capabilities, thereby improving the model's overall reasoning performance.

The contributions of this study are threefold. First, we introduce a novel slow-thinking inference strategy, Atomic Reasoner, aiming to simulate o1-like slow-thinking reasoning without training. Secondly, we define a set of abstract, broadly applicable atomic reasoning actions and incorporate a cognitive routing mechanism to provide fine-grained guidance during reasoning, thereby enhancing inference efficiency. 
Lastly, AR demonstrates substantial performance improvements across several benchmarks for logical reasoning. Notably, Atomic Reasoner exhibits strong scalability; by extending the set of atomic actions or the routing mechanism, AR can be well-adapted to downstream tasks, to effectively improving reasoning quality without incurring significant inference costs.

\section{Related Work}
The enhancement of reasoning capabilities in LLMs has garnered significant research attention, with approaches generally categorized into two types based on whether training is involved.
\subsection{Training Scaling-Driven Reasoning Enhancement}
This category optimizes the training data, architecture, or algorithms, resulting in a new LLM with enhanced reasoning abilities.
\paragraph{High-quality Reasoning Data Synthesis and Supervised Fine-tuning~(SFT)}
SFT is the foundational paradigm for LLM training. Collecting high-quality reasoning data, the models are fine-tuned to align with the distribution of corpora~\cite{chung2024scaling,zhou2024lima,zelikman2022star,guan2025rstar,zhang2024rest}. However, these methods remain constrained by the cost of acquiring high-quality data.

\paragraph{Optimization-Based Model Fine-tuning}
Another training approach involves RL~\cite{sutton2018reinforcement} optimization guided by human preference (a.k.a, RLHF~\cite{ziegler2019fine,yuan2024uni}), AI preference~\cite{liu2024enhancing}, or appropriate reward signals~\cite{gao2024designing,trung2024reft,zhai2024fine,jaech2024openai}. RL drives LLMs to maximize cumulative rewards, fostering the emergence of high-quality slow-thinking reasoning capabilities. Recent studies demonstrate that RL alone can significantly enhance LLM reasoning (e.g., R1-Zero~\cite{guo2025deepseekr1}). While these methods do not rely on data synthesis and have higher performance ceilings, they require sophisticated training mechanism designs and substantial computational resources.

\subsection{Test-Time Inference Scaling-Driven Reasoning Enhancement}
This category enhances model performance during inference without further training. The basic scaling method is to decompose complex problems into multiple sequential subtasks~\cite{wei2022cot}.
This often combine horizontal scaling strategies (e.g., SC-CoT~\cite{0002WSLCNCZ23sccot}) to generate multiple solutions in parallel for optimal selection.
When dealing with larger solution spaces, search mechanism has been introduced for better solution chains, including tree-structured search with necessary pruning~\cite{yao2024tot} or Monte Carlo Tree Search (MCTS) for iterative node expansion and backpropagation~\cite{jiang2024technical}. Node quality assessment is crucial in search process, with some methods using LLMs as verifiers~\cite{li2024generation} while others train specialized reward models for specific tasks (e.g., mathematics)~\cite{ma2023let,luo2024improve,kang2024mindstar}. These approaches heavily depend on verifier accuracy and also face computational cost challenges.

We also acknowledge that guiding structured thinking is beneficial for LLM reasoning. However, in contrast to previous tree-based approaches, we present that introducing finer-grained and systematic guidance in reasoning process is crucial for reducing computational overhead. This can also improve the logical coherence and effectiveness of O1-like slow-thinking reasoning process.

\subsection{Reproductions Inspired by O1}
Efforts to reproduce O1-like slow-thinking reasoning have emerged, forming two distinct approaches as mentioned above. The first category focuses on training, involving SFT by generating slow-thinking corpus using methods such as Monte Carlo Tree Search (MCTS)~\cite{zhao2024marco}, or leveraging RL to optimize model parameters guided by specific reward signals~\cite{wang2024openr,zhang2024o1coder}. The second category, in contrast, aims to achieve O1-like slow-thinking reasoning without fine-tuning models, relying instead on the design of reasoning frameworks. Examples include guiding self-reflection and step-by-step reasoning through in-context learning (ICL)~\cite{g1}, integrating MCTS into the reasoning and planning process~\cite{jiang2024technical,yanplanning,zhang2024llama,llamaberry}, or constructing reasoning trees via depth-first search (DFS) for high-quality data generation~\cite{o1journey}.
Unlike existing work, we aim to avoid the need for large-scale search, and emphasize a finer-grained , more structured cognitive guidance through aqtomic reasoning operations, to foster a more systematic and deliberate thought process during inference.
\begin{figure}[t]
\centering
  \includegraphics[width=0.95\linewidth]{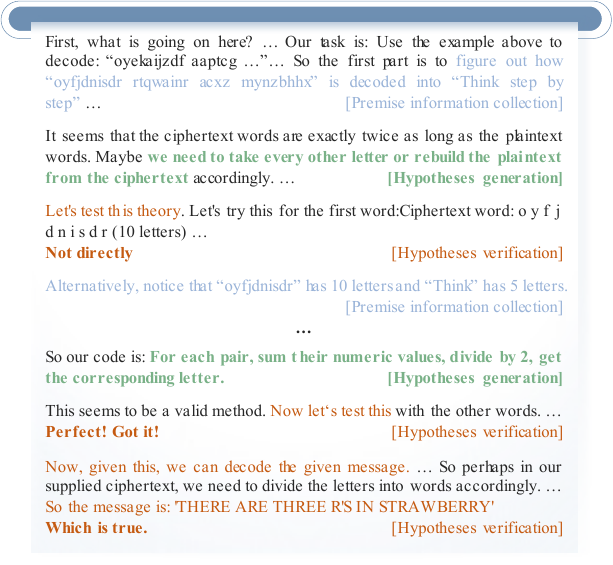}
  \caption {Illustration of the analysis of the slow thinking process for o1, with the case from the OpenAI website.}
  \label{fig:analysis}
\end{figure}

\section{Atomic Reasoner}

This study introduces a novel reasoning framework called Atomic Reasoner (AR), designed to emulate O1-like slow thinking through structured and systematic reasoning processes. The core of AR lies in atomizing the reasoning process into several granular steps, thereby providing more precise guidance and reducing the cognitive complexity of LLMs during reasoning. Additionally, AR introduces a tree-based atomic cognitive routing mechanism, enabling reasoning that is more structured, logically rigorous, and cognitively robust. 

\subsection{Case Study: Analysis of Behavioral Patterns in slow-thinking process}
\label{sec:atomic_ana}
Existing cognitive theory research indicates that the slow-thinking process involves deliberate, effortful, and analytical reasoning~\citet{kahneman2011thinking}. To explore the behavioral patterns associated with this process, we conduct a qualitative case study based on the output examples published on the official website of O1 model~\cite{o1}. Figure~\ref{fig:analysis} provides an illustrative analysis case. We identify three core behavioral patterns as follows:

\textbf{\circledspace{1}Decomposition of the problem}: The process breaks the problem down into smaller, manageable sub-steps, with fully execution.

\textbf{\circledspace{2}Thorough comprehension and organization of the premise information}: Many steps in examples involve restating or extracting deeper insights about premise information, thereby enriching the information for subsequent reasoning.

\textbf{\circledspace{3}Continuous hypothesis formulation and rigorous verification checks}: The hypotheses are not only executed but also undergo thorough validation to ensure their accuracy and relevance.

\begin{figure*}[t]
\centering
  \includegraphics[width=0.89\linewidth]{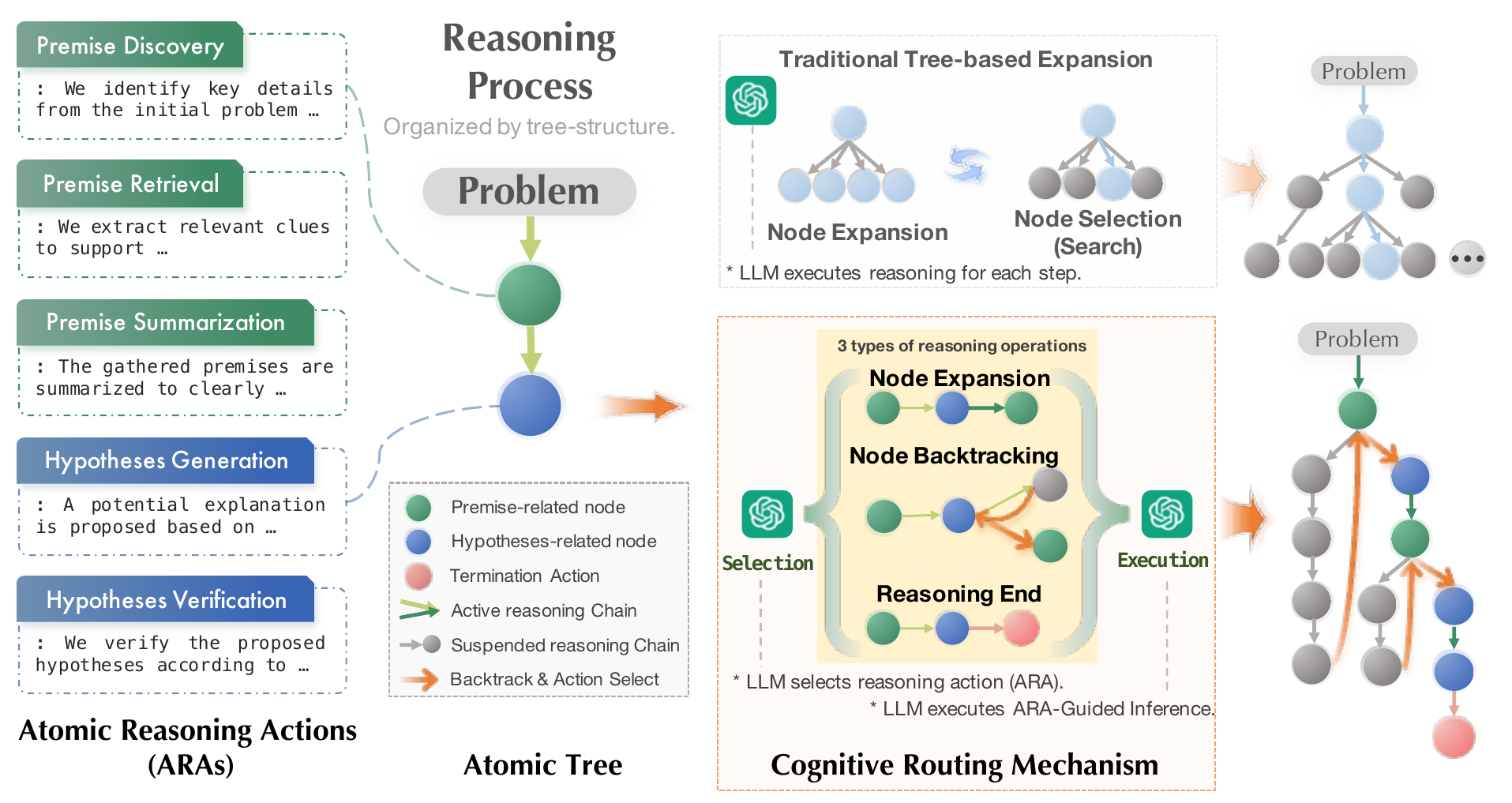}
  \vspace{-6px}
  \caption {Overall architecture of Atomic Reasoner (AR). AR introduces Atomic Reasoning Actions and proposes an Atomic Tree-based organization of reasoning processes. Unlike traditional Tree-Based Expansion methods (top right) that rely on LLM's parallel node expansion and complex selections, AR adopts a cognitive reasoning approach, guiding LLMs to first select appropriate reasoning actions, followed by precise operations (node expansion, backtracking, or termination) on the Atomic Tree, ultimately achieving more efficient deliberate reasoning.}
  \label{fig:framework}
\end{figure*}

Noting that \citet{wu2024comparative} proposed that self-reflection and decomposition are the two main behavioral patterns in O1, which aligns with our findings. 
Compared to the intricate trial-and-error process, this cognitive thinking, which decomposes actions into atomic steps and executes them sequentially, demonstrates higher structural coherence and clarity, better aligning with the human cognitive paradigm of problem decomposition and stepwise resolution.
These patterns inspire us to define refined, atomic-level reasoning actions~(ARAs), denoted as $\mathbf{\Lambda} = \{ a_1, a_2, a_3, \ldots, a_m \}$, where $a_i$ represents a distinct action. By executing a well-structured and logical sequence of these actions, we can guide LLMs to construct complex, deliberate reasoning processes. 
Next, we will discuss why this approach is effective and provide a rationale for the design scheme.
We argue that:
\begin{remark}
At the same thought scale, reasoning guided by ARAs can effectively reduce entropy, minimizing the cognitive uncertainty in the LLMs output and improving the overall reasoning efficiency.
\end{remark}
\noindent \textbf{Analysis.}  
We analyze the advantages of ARAs in slow-thinking reasoning from the perspective of entropy. For the sake of analytical tractability, we assume uniform and steady changes in information entropy throughout the reasoning steps. 
Assume that traditionally, a problem is decomposed into $N$ independent reasoning steps. The entropy of each step $i$ can be expressed as:
\begin{equation}
H_i = -\int_{R_i} p_i(x) \log p_i(x) \, dx
\end{equation}
where $p_i(x)$ represents the probability density function of different outcomes at step $i$, and $R_i$ is the space of outcomes for this step.

Next, we consider guiding the reasoning process using ARAs ($\mathbf{\Lambda}$ with $M$ actions). Within this framework, the problem is still divided into $N$ steps, but each step's reasoning is executed by selecting an atomic reasoning action. The entropy under each reasoning action $a_j$ is given by:
\begin{equation}
E_{a_j} = -\int_{S_{a_j}} q_{a_j}(y) \log q_{a_j}(y) \, dy
\end{equation}
where $q_{a_j}(y)$ is the probability density function for action $v$, and $S_{a_j}$ is its outcome space.
The entropy of each step is represented by the weighted entropy of the chosen atomic reasoning actions:
\begin{equation}
\label{eq:entropy_atomic}
H'_i = \sum_{j=1}^{M} r_{i,j} E_{a_j}
\end{equation}
where $r_{i,j}$ is the probability of selecting the $j$-th atomic reasoning action ${a_j}$ in the $i$-th step.

Compared to reasoning directly in the original space, the ARA constrains the execution of each sub-step within a smaller solution space ($S_{a_j}$). This means that within each atomic action, the probability distribution $q_{a_j}(y)$ tends to be more concentrated due to its specialized design. 
The smaller reasoning subspace and the more focused distribution result in lower entropy~($E_{a_j}$) for the output of the LLM under each atomic action. 

In contrast, the probability distribution $p_i(x)$ in traditional methods is often more dispersed, resulting in higher entropy compared to the weighted average of the entropies $E_{a_j}$ associated with ARAs, i.e., $H_i > H'_i$. This indicates that ARAs reduce the cognitive uncertainty with great potential at each reasoning step, thereby making the overall reasoning process more efficient and accurate. \hfill $\square$

Building upon the analysis of the reasoning pattern and the effectiveness of ARAs, two key questains arise: how to define ARAs, and how to implement a rational ARA selection distribution $(r_{i,j})$ to make the overall entropy more regulated. These will be discussed in detail in \S~\ref{sec:atomic_def} and \S~\ref{sec:atomic_route}, respectively.

\subsection{The Atomic Definition of Reasoning Processes}
\label{sec:atomic_def}

The analysis in \S~\ref{sec:atomic_ana} presents the effectiveness of atomic reasoning actions (ARAs). In the following, we discuss the definition of the set of ARAs. 
O1 reasoning pattern reveals that the extraction and utilization of premise information consistently play a pivotal role in maintaining logical continuity throughout the reasoning process. Thus, we categorize ARAs into two distinct types: premises-related and reasoning-related actions.

\paragraph{Premise-related ARAs}
This includes \textit{premise discovery}, \textit{premise retrieval}, and \textit{premise summarization}. They are designed to assist the model in systematically collecting and organizing closely relevant premise knowledge. For instance, at the very beginning, the model employs ``premise discovery'' to identify explicit clues and their implicit implications, and then ``premise retrieval'' enables the acquisition of specific clues and premise information pertinent to each reasoning step.

\paragraph{Reasoning-related ARAs}
This pertains to actions that involve appropriate \textit{hypotheses generation} and \textit{hypotheses verification} at each step of reasoning. These actions work in concert to jointly ensure the reliability and accuracy of the reasoning process. ``Hypothesis verification'' must be executed promptly after a hypothesis is proposed. Also, if an existing hypothesis is found to be flawed, the model must adapt by formulating a new hypothesis, maintaining the coherence and logical consistency of the reasoning chain.

The definition of ARAs clarifies the key patterns of the slow-thinking reasoning, and emphasizes the importance of thoroughly observing the problem itself. This also aligns with the philosophical cognition on thought process~\cite{dewey1997we}.

\subsection{Cognitive Routing for Atomic Reasoning}
\label{sec:atomic_route}

Building upon ARAs, we need to generate rational action selection distributions, ensuring to reduce entropy of reasoning. 
Tree search achieves this through exhaustive exploration, with its distinctive advantages being structured representation. However, it relies on validators and involves computational overhead. To address this challenge, we formalize {Atomic Trees} to organize the reasoning process, clearly representing logical relationships within the reasoning structure. Additionally, we propose a {Cognitive Routing Mechanism} that operates on this complex tree through more fine-grained guidance based on ARAs. This mechanism more closely resembles human thinking patterns - relying on intuition and experience rather than bottom-up search - enabling more efficient node expansion and thus more effective construction of complete, high-quality reasoning paths.

\paragraph{Atomic Tree} Slow-thinking reasoning often involve the exploration and manipulation of multiple reasoning chains. To better manage this complexity, we introduce the Atomic Tree. Each node in this tree corresponds to a specific reasoning action and contains its execution details. The chains in the tree represent linear extensions of the reasoning path, while branches indicate the bifurcation of reasoning at specific nodes into multiple pathways. The Atomic Tree provides a structured and logically transparent representation of the reasoning process, enabling LLMs to better understand and navigate through the process.

\paragraph{Cognitive Routing Mechanism}

As shown in Figure~\ref{fig:framework}, the reasoning process involves three core operations with the atomic tree: extending reasoning chains, expanding branches following backtracking, and termination. Specifically, the LLM decides which reasoning action to pursue at the current state. When the LLM determines a reasoning chain is complete and challenging to extend further, it backtracks to a historical node on the current chain and initiates a new action, to expand a new branch. As the reasoning progresses via cognitive routing, the Atomic Tree grows continuously until termination—either actively when a solution is reached or passively due to scale limits. Ultimately, the LLM derives the answer from the expanded reasoning process.

In this process, reasoning chains exist in three status: (1) {active}, currently being explored; (2) {suspended}, completed and not extended further; and (3) {dormant}, awaiting branching. 
Together, based on a foundational structure—the Atomic Tree—we introduce two core operations, resulting in three distinct states for multiple reasoning chains. These collectively form the {Cognitive Routing Mechanism}, enabling systematic and efficient reasoning.

\begin{figure}[t]
  \includegraphics[width=0.96\columnwidth]{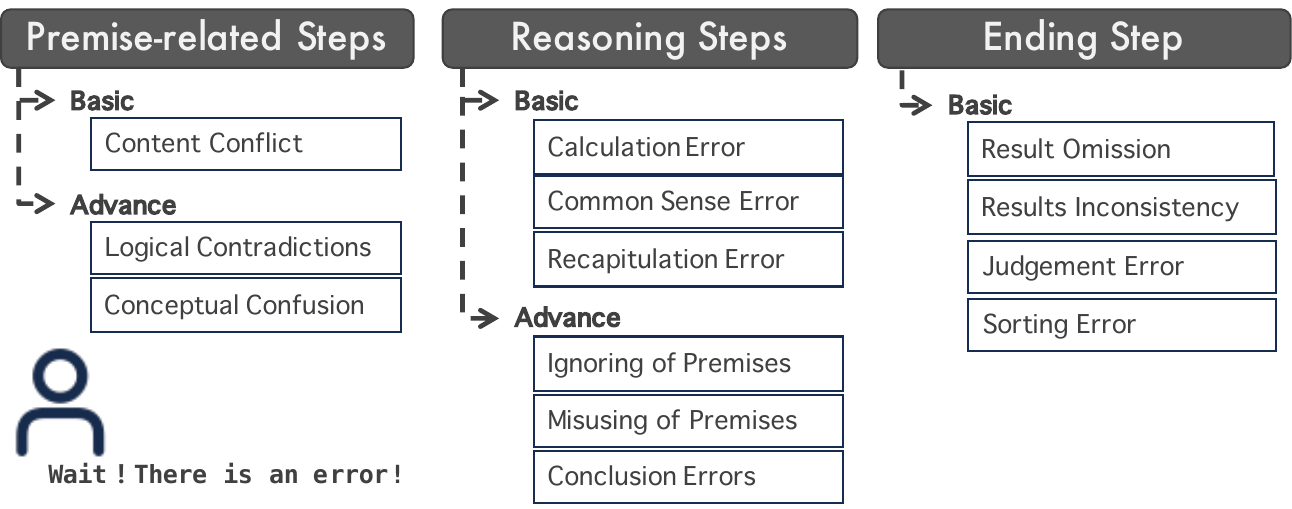}
  \caption{Schematic illustration of error types defined in the Checker mechanism of AR.}
  \label{fig:checker}
\end{figure}

\subsection{Advanced Reliability  Enhancement}
To enhance the efficiency and accuracy of reasoning, several optimizations are introduced.

\paragraph{Domain-Specific Adaptation through SOPs}
Reasoning tasks often exhibit specific patterns depending on their domains. To achieve this, we introduce Standard Operating Procedures (SOPs), domain-specific adaptations to optimize reasoning. As shown in Figure~\ref{fig:sop}, each SOP consists of strategies for executing ARA, scheduling strategies for reasoning factors, and example-based guidance. At the initial triage stage, problems are classified and routed to the most appropriate SOP. 
This enables the AR framework to re-use human expert experience, maximizing atomic reasoning's potential for better efficiency and accuracy.

\paragraph{Fine-Grained Reflection and Self-improving}
To ensure the effectiveness of the reflection and optimization process, we designed a fine-grained checker mechanism for different types of atomic actions. Figure~\ref{fig:checker} shows error types that with specific rules defined for each type of atomic action to identify and correct, the checker ensures the effectiveness of the reflection and optimization process.

\begin{table*}[t]
\renewcommand{\arraystretch}{1.0}
\centering
\small
\begin{tabular}{>{\raggedright}m{0.12cm}>{\raggedright}m{3.6cm}|>{\raggedleft}m{0.9cm}>{\raggedleft}m{0.8cm}>{\raggedleft}m{1.0cm}>{\raggedleft}m{0.9cm}>{\raggedleft}m{0.75cm}>{\raggedleft}m{0.9cm}>{\raggedleft}m{0.8cm}>{\raggedleft}m{0.7cm}>{\raggedleft}m{0.7cm}}
\toprule
\rowcolor[HTML]{F5FDFF} 
\cellcolor[HTML]{F5FDFF} &
  \cellcolor[HTML]{F5FDFF} &
  \cellcolor[HTML]{F5FDFF} &
  \cellcolor[HTML]{F5FDFF} &
  \cellcolor[HTML]{F5FDFF} &
  \multicolumn{3}{c}{\cellcolor[HTML]{F5FDFF}\textbf{ZebraGrid}} &
  \multicolumn{3}{c}{\cellcolor[HTML]{F5FDFF}\textbf{BBH}} \\
\rowcolor[HTML]{F5FDFF} 
\multirow{-2}{*}{\cellcolor[HTML]{F5FDFF}\textbf{}} &
  \multirow{-2}{*}{\cellcolor[HTML]{F5FDFF}\textbf{Method}} &
  \multirow{-2}[3]{*}{\cellcolor[HTML]{F5FDFF}\textbf{AIME}} &
  \multirow{-2}[3]{*}{\cellcolor[HTML]{F5FDFF}\textbf{MATH}} &
  \multirow{-2}[3]{*}{\cellcolor[HTML]{F5FDFF}\textbf{TheoremQA}} &
  \textbf{easy} &
  \textbf{hard} &
  \textbf{avg.} &
  \textbf{ws.} &
  \textbf{cj.} &
  \textbf{avg.} \tabularnewline
\midrule
 & CoT~\cite{wei2022cot} & 2.67 & \textbf{52.9} & 33.4 & 68.4 & 9.09 & 28.9 & 50.0 & 36.4 & 45.1  \tabularnewline
 & SC-CoT~\cite{0002WSLCNCZ23sccot} & \textbf{4.3} & 50.1 & \textbf{36.0} & \underline{\textbf{68.9}} & 5.5 & 28.5 & \underline{\textbf{52.9}} & 36.4 &  \underline{\textbf{46.5}}\tabularnewline
 & ToT~\cite{yao2024tot} & 2.67 & \underline{\textbf{52.5}} & 25.7 & 27.1 & 25.1 & 25.8 & 47.1 & 26.1 & 37.7  \tabularnewline
 & BoN & 0.7 & 48.8 & 27.3 & 67.9 & 12.4 & 30.9 & 49.0 & 35.8 & 43.4   \tabularnewline
 & G1~\cite{g1} & 0.0 & 5.7 & 11.8 & 60.2 & 13.3 & 29.0  & 0.0 & 18.2 & 16.2   \tabularnewline
 & Llamaberry~\cite{llamaberry} & \underline{\textbf{4.0}} & 49.8 & 28.0  & 66.5 & \underline{\textbf{28.4}} & \underline{\textbf{41.1}} & 37.0  & \underline{\textbf{46.1}} & 29.1 \tabularnewline
 & Plan-MCTS~\cite{yanplanning} & 2.67 & 38.7 & 21.2 & 57.3 & 24.8 & 35.3 & 44.1 & 35.1 & 43.4  \tabularnewline
 \cmidrule(l){2-11} 
\multirow{-8.5}{*}{\rotatebox{90}{GLM-4-FlashX}} & 
\cellcolor{cyan!10}\textbf{Atomic Reasoning~(AR)} & 
\cellcolor{cyan!10}\underline{\textbf{4.0}} & 
\cellcolor{cyan!10}\textbf{52.9} & 
\cellcolor{cyan!10}\underline{\textbf{33.9}} & 
\cellcolor{cyan!10}\textbf{69.8} & 
\cellcolor{cyan!10}\textbf{33.7} & 
\cellcolor{cyan!10}\textbf{44.3} & 
\cellcolor{cyan!10}\textbf{60.1} & 
\cellcolor{cyan!10}\textbf{47.1} & 
\cellcolor{cyan!10}\textbf{52.0} \tabularnewline
\midrule \midrule
 & CoT~\cite{wei2022cot} & 7.33 & 76.8 & 54.9 &  84.9 & 40.7  & 52.3 & 77.9 & 55.5 & 65.0  \tabularnewline
 & SC-CoT~\cite{0002WSLCNCZ23sccot} & 8.0 & 77.7& 52.5 & 83.6 & 29.7  & 47.7 & \underline{\textbf{85.3}} & 56.4 & 67.7 \tabularnewline
 & ToT~\cite{yao2024tot} & 11.3 & 78.1 & 34.7 &  79.5 & 37.0  & 51.0 & 66.7 & 51.5 &60.3   \tabularnewline
 & BoN & \textbf{18.0} & \textbf{81.5} & 57.6 &  \underline{\textbf{85.5}} & \underline{\textbf{43.8}}  & \underline{\textbf{56.6}} & 76.5 & 50.9 & 67.7  \tabularnewline
 & G1~\cite{g1} & 8.67 & 70.1 & 55.8 & 82.5 & 37.6  &52.7 & 83.3 & 55.8 & 68.0    \tabularnewline
 & Llamaberry~\cite{llamaberry} & \underline{\textbf{13.3}} & \underline{\textbf{80.1}} & \underline{\textbf{58.2}}  &  85.2 & 39.9  & 55.1 &  82.4 & \underline{\textbf{57.6}} & \underline{\textbf{69.0}}  \tabularnewline
 & Plan-MCTS~\cite{yanplanning} & 9.0 & 79.1 & 55.9 &  80.0 &  38.7 & 52.0 & 82.4 & 56.4& 68.7  \tabularnewline
 \cmidrule(l){2-11} 
\multirow{-8.5}{*}{\rotatebox{90}{GPT-4o-mini}} & 
\cellcolor{cyan!10}\textbf{Atomic Reasoning~(AR)} & 
\cellcolor{cyan!10}\underline{\textbf{13.3}} & 
\cellcolor{cyan!10}{79.1} & 
\cellcolor{cyan!10}\textbf{59.6}& 
\cellcolor{cyan!10}\textbf{86.5} & 
\cellcolor{cyan!10}\textbf{45.3} & 
\cellcolor{cyan!10}\textbf{58.7} & 
\cellcolor{cyan!10}\textbf{89.2} & 
\cellcolor{cyan!10}\textbf{75.2} & 
\cellcolor{cyan!10}\textbf{78.8}\tabularnewline
\bottomrule
\end{tabular}
\caption{Performance results showing the mean values of 3 trials across all test sets. The highest values for each group of results are shown in bold, and the second-highest values are shown in bold with underline.}
\label{tab:tab1_results}
\end{table*}

\section{Experiment}
Our experiments were designed to validate three research questions (RQs):

\textbf{RQ1}: What tasks is AR good at, guiding LLM to engage in slow thinking and improve performance?

\textbf{RQ2}: How do different modules of AR affect its performance?

\textbf{RQ3}: Whether AR's output demonstrate improved reasoning behaviors in complex scenarios?

\subsection{Experimental Setup}
To demonstrate the value of the AR framework, we select well-known LLMs GPT-4o-mini and GLM-4-flashX for validation. These models are accessed via APIs provided by infrastructure platforms such as OpenAI Platform and BigModel. 

The datasets used for model validation are divided into two categories: mathematical and scientific reasoning problems and linguistic logic reasoning problems (including BBH and ZebraGrid). To minimize resource consumption during testing, we randomly selected subsets of questions from each dataset to form a test repository. We test 3-trials for each evaluation and report the mean values in the following, thereby mitigating statistical errors. Evaluation metrics primarily included task completion success rates.

To demonstrate the advantage of AR in controlling resource overhead, we have set the maximum inference rounds during the AR inference process to 12. This significantly reduces the overhead compared to traditional tree search methods. In Appendix C, we provide a detailed account of the experimental setup, including the implementation and parameter settings of baseline algorithms, as well as the construction of the test set. To enhance the reproducibility of our experiments, we have included portions of the code in the supplementary materials for review. This includes the code implementation of the AR inference process and information on the randomly constructed test set.


\subsection{RQ1: Performance of the AR Framework in Complex Logical Reasoning Tasks}

We evaluated the capability of the AR framework to guide LLMs in handling complex logical reasoning tasks. Through comprehensive testing across various datasets, as presented in Table~\ref{tab:tab1_results}, we observed that AR significantly enhances reasoning performance in logical reasoning tasks, particularly in linguistic inference problems. Specifically, for linguistic logic reasoning tasks, the AR-enhanced inference consistently outperformed all other strategies, especially in hard tasks, demonstrating a substantial lead. This indicates AR's powerful role in enhancing LLMs' capacity for complex, deliberative logical analysis and reasoning.

However, mathematical tasks remain challenging. While AR consistently exhibits performance levels exceeding those of CoT, the advantage is not as pronounced as other tasks. Notably, in mathematical tasks, negative optimization phenomena occur more frequently in baseline algorithms; for instance, with GLM-4-FlashX on math tasks, almost all algorithms underperformed compared to CoT. This highlights the unique challenges of mathematical tasks where, without specialized external feedback signals (such as verifier models or PRM, which were not included in these strategies), LLMs are constrained by their inherent knowledge systems and struggle to achieve significant improvements through multi-round thinking alone. Nevertheless, our method demonstrated consistent stability across all scenarios without severe performance fluctuations in specific cases, validating that the AR reasoning framework can achieve robust and stable reasoning through refined cognitive routing mechanisms.
\begin{figure}[t]
\centering
  \includegraphics[width=0.24\textwidth]{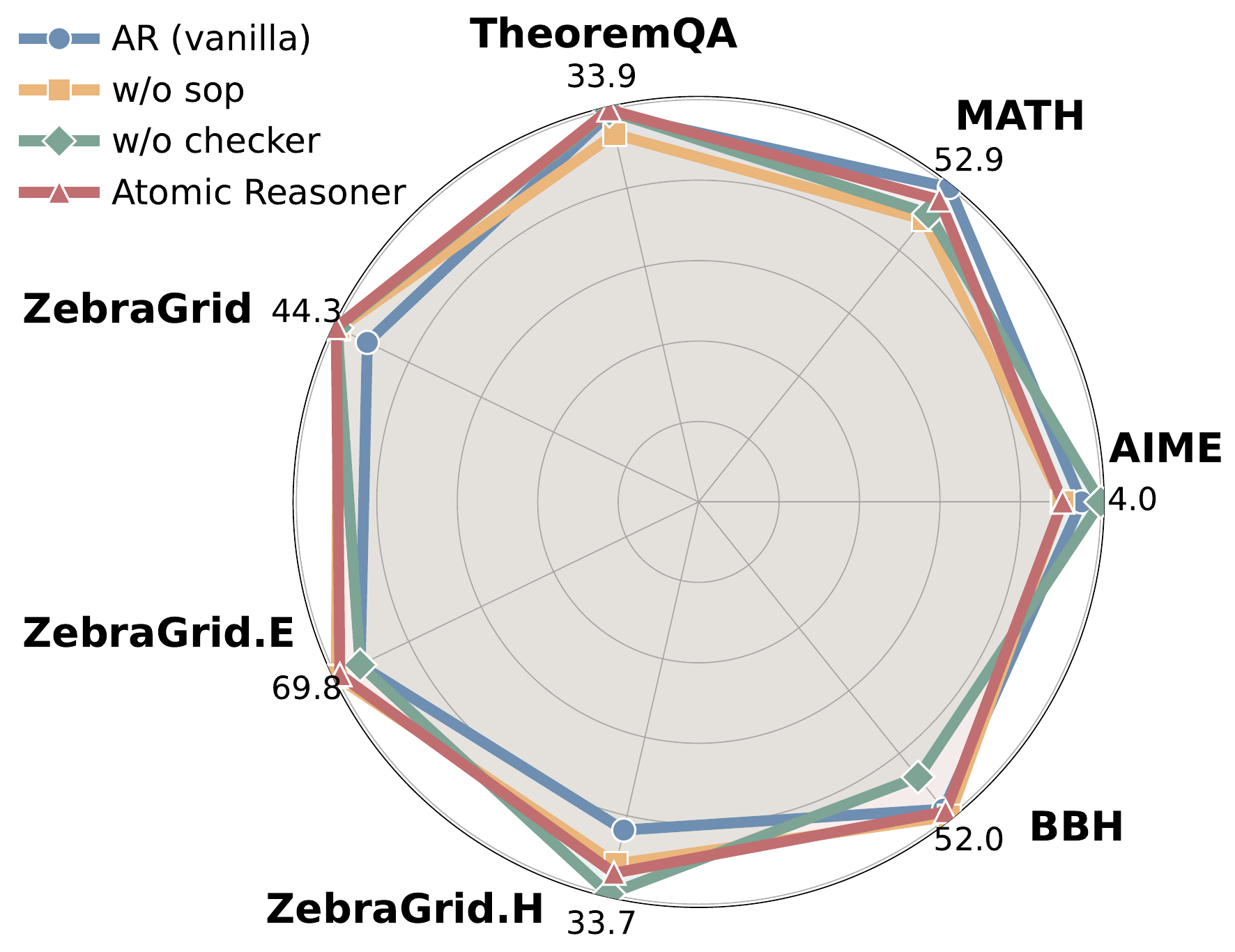}\hspace{-4px}
  \includegraphics[width=0.24\textwidth]{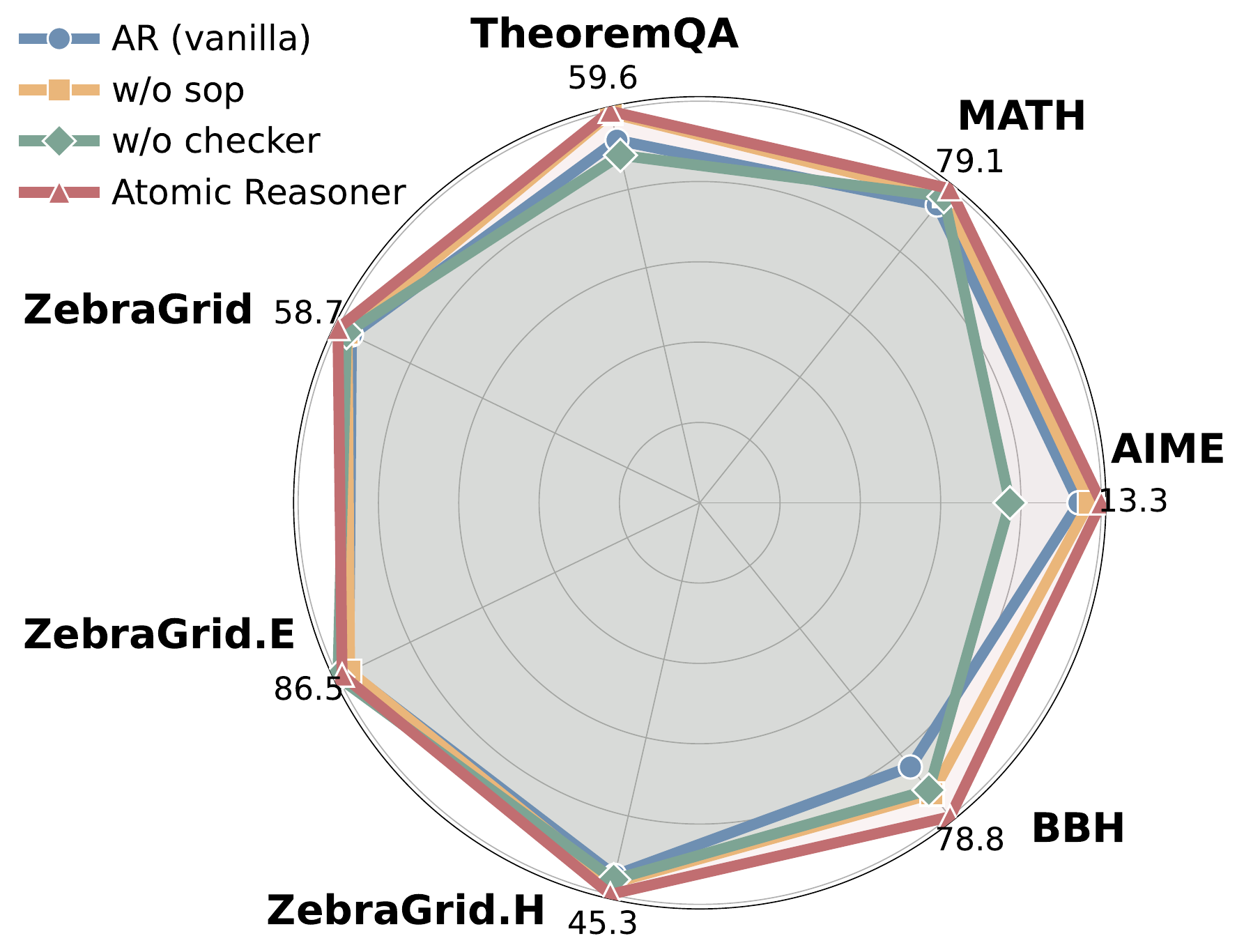}
  \caption{Ablation studies on different mechanisms in AR, with results from model GLM-4-flashX (left) and model GPT-4o-mini (right).}
  \label{fig:ablation}
\end{figure}

\begin{figure*}[t]
  \includegraphics[width=\textwidth]{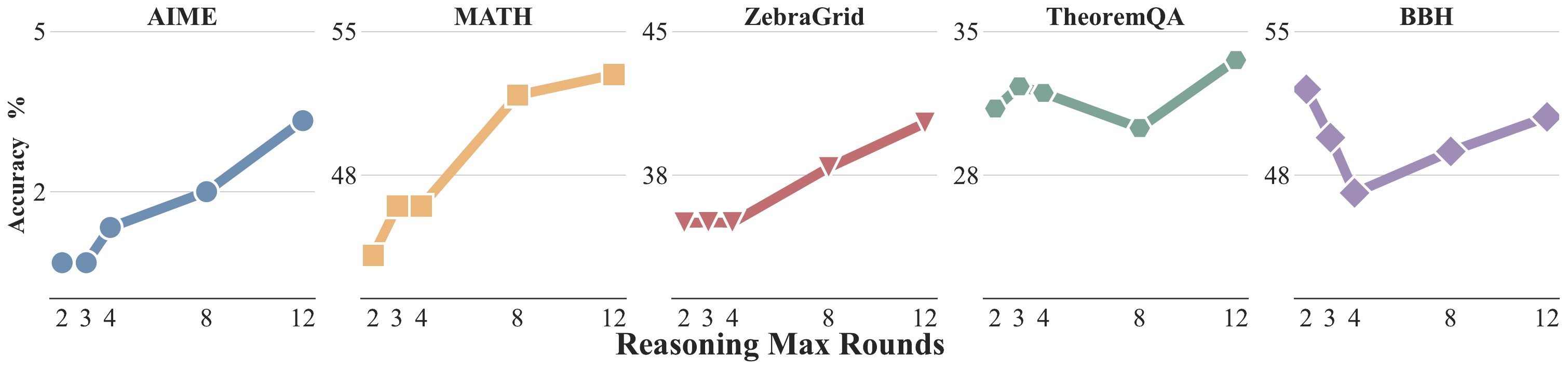}
  \caption{Performance curves with respect to the scaling of Maximum Reasoning Rounds.}
  \label{fig:scaling}
\end{figure*}
\begin{table*}[]
\centering
\small
\begin{tabular}{>{\raggedright}m{4.25cm}>{\raggedright}m{2.0cm}|>{\raggedleft}m{1.1cm}>{\raggedleft}m{1.25cm}>{\raggedleft}m{1.5cm}>{\raggedleft}m{1.5cm}>{\raggedleft}m{1.0cm}}

\toprule
\textbf{Model}                & \textbf{Method}        & \textbf{AIME} & \textbf{MATH} & \textbf{TheoremQA} & \textbf{ZebraGrid}      & \textbf{BBH}           \tabularnewline \midrule
Qwen2.5-1.5b-Instruct            & CoT                    & 5.0          & 58.0          & 26.9               & 17.99          & 9          \tabularnewline
\textbf{Qwen2.5-1.5b-Instruct}   & \textbf{FT w/ AR data} & \textbf{8.0}          & 51.5             & \textbf{31.4}       & \textbf{33.8} & \textbf{22.7} \tabularnewline
\midrule
DeepSeek-R1-Distill-Qwen-1.5B & \textbf{-}             & 12.7         & 67.3         & 37.0               & 13.2           & 17.2         \tabularnewline \bottomrule
\end{tabular}
\caption{Fine-tuning results of Qwen2.5-1.5b-Instruct model using AR-generated data, compared with performance of Qwen2.5-1.5b-Instruct and DeepSeek-R1-Distill-Qwen-1.5B.}
\label{tab:finetuning}
\end{table*}

\subsection{RQ2: Ablation Study}

\paragraph{SOP and Checker Mechanisms}
To assess the contributions of individual components within the AR framework, we conducted ablation studies by removing key mechanisms such as SOP and Checker, as shown in Fig.~\ref{fig:ablation}. Results indicate that without the SOP module, different base models exhibit inconsistent performance degradation across various datasets. This suggests two key findings: firstly, our defined SOP process is well-founded and contributes to stable reasoning; secondly, the potential for enhanced performance exists through more sophisticated experience injection via SOP.

When the checker module is disabled, performance across most tasks is notably impacted. This validates our hypothesis that the checker mechanism, through its well-defined structure, can effectively identify certain errors as intended. These findings demonstrate that both the SOP and checker components play crucial roles in maintaining the system's overall performance and reliability.

\paragraph{Impact of Maximum Reasoning Rounds Scaling}
To evaluate the influence of maximum reasoning rounds on performance, we modified this hyper-parameter from its default value of 12 to various settings (2, 3, 4, and 8). To eliminate confounding factors, this verification was conducted on vanilla AR, excluding additional modules such as SOP and checker, and solely incorporating atomic reasoning definitions and cognitive routing mechanisms. The demonstrate a general trend of performance improvement with increasing maximum reasoning rounds. An notable exception occurs in BBH, where peak performance was achieved with 2 rounds used, suggesting that only a single premise-discovery operation likely occurred. This can be reasonably attributed to the nature of BBH tasks, which may not require complex multi-step deliberative thinking. However, the performance degradation in BBH was quickly contained, and performance recovered with additional rounds, further demonstrating the robustness of  our strategy.

\subsection{RQ3: Effectiveness of Data Synthesized by AR}

To demonstrate AR's effectiveness in deliberative reasoning, we present two cases in the appendix that showcase the framework's inference process. These examples illustrate rational cognitive behaviors including comprehensive premise collection, iterative hypothesis generation and validation, self-examination, and thought exploration through node backtracking and branching paths.

We also explored AR's potential in data synthesis for LLM fine-tuning. Models fine-tuned with AR-synthesized data showed significant performance improvements, with some tasks achieving nearly double the effectiveness of the baseline model (Tab.~\ref{tab:finetuning}). We also report the results on DeepSeek-R1-Distill-Qwen-1.5B, which is distilled from the DeepSeek-R1 on a 1.5B model, further validating the improvements achieved through fine-tuning with our collected deliberative thinking data. This indicates that the high-quality deliberative reasoning data generated by the AR framework holds substantial value for model training.

It should be noted that model fine-tuning and data synthesis were not the primary focus of this research; these experiments were conducted merely to validate ARs viability. While we did not conduct larger-scale experiments, we will open-source our code, confident that AR will contribute to high-quality data synthesis and SFT in the community.

\section{Conclusion}
This paper presents Atomic Reasoner (AR), a novel framework that enhances LLMs' slow-thinking capabilities through atomic reasoning actions and cognitive routing mechanisms. By decomposing complex reasoning into fundamental atomic steps and providing fine-grained guidance, AR demonstrates significant improvements in reasoning performance . Our experimental results show both immediate performance gains and the potential for model improvement through generated high-quality reasoning data. The framework's modular design and extensibility make it particularly valuable for practical applications across different domains. These findings suggest that structured, atomic-level approaches to reasoning is a promising direction for advancing AI systems' capability.

\section*{Limitations}
The proposed Atomic Reasoner (AR) framework demonstrates significant advantages in complex reasoning tasks through its structured and systematic approach to enhancing LLMs' reasoning capabilities. However, several limitations warrant consideration. First, the framework encounters challenges in handling highly dynamic and uncertain tasks, particularly where test-time computation adversely affects user experience. Second, while Standard Operating Procedures (SOPs) have improved domain adaptability, the framework's current implementation only provides general applicability and extensibility, rather than the comprehensive, domain-specific customization often required by diverse application scenarios.

Future research directions include the integration of reinforcement learning and fine-tuning techniques to enable adaptive optimization of reasoning strategies, thereby addressing a broader spectrum of complex tasks. The ultimate objective is to develop a more sophisticated reasoning system that maintains high performance and reliability across diverse domains while offering enhanced flexibility and computational efficiency.

\bibliography{main}

\clearpage

\appendix
\section{Discussion: The Relationship Between Atomic Reasoning Actions and Dewey's Five Steps of Reflective Thinking}
John Dewey's Five Steps of Reflective Thinking~\cite{dewey1997we}—comprising the recognition of a felt difficulty, the precise location and definition of the problem, the suggestion of possible solutions, the reasoning about the implications of these suggestions, and further observation and experimentation to accept or reject them—embody a profound philosophical significance. Dewey's framework emphasizes a systematic and reflective approach to problem-solving, underscoring the importance of critical thinking and empirical validation.

The initial stages of Dewey's model, which involve identifying difficulties and defining problems through the systematic collection of relevant information, serve as a benchmark of philosophical rigor. Our approach, which aligns with these premise-related actions, reflects a similar commitment to thorough understanding and articulation of the problem at hand. This alignment with Dewey's meticulous methodology indirectly underscores the validity and soundness of our framework. Subsequently, the reasoning-related actions, particularly the generation and verification of hypotheses, mirror Dewey's emphasis on exploring potential solutions and evaluating their validity through reasoned analysis and empirical testing.

The congruence between Atomic Reasoning Actions (ARAs) and Dewey's Five Steps of Reflective Thinking underscores the logical and well-founded nature of our definition of atomic reasoning. Our conclusion through analysis, fortuitously aligns with Dewey's systematic and reflective thinking framework. This alignment serves to validate the soundness of our analysis and the definitions we have established. By integrating Dewey's philosophical principles with the structured decomposition of reasoning processes, ARAs offer a robust framework for addressing complex problems. This integration not only preserves the philosophical depth of Dewey's approach but also enhances its applicability in contemporary contexts, where precision and efficiency in reasoning are paramount. Thus, the definition of atomic reasoning actions is justified, as it bridges traditional philosophical principles with modern computational methods, providing a comprehensive and practical approach to problem-solving.

\section{Implementation Details}
\subsection{Design Details about SOP}

\begin{figure*}[t]
\centering
  \includegraphics[width=2\columnwidth]{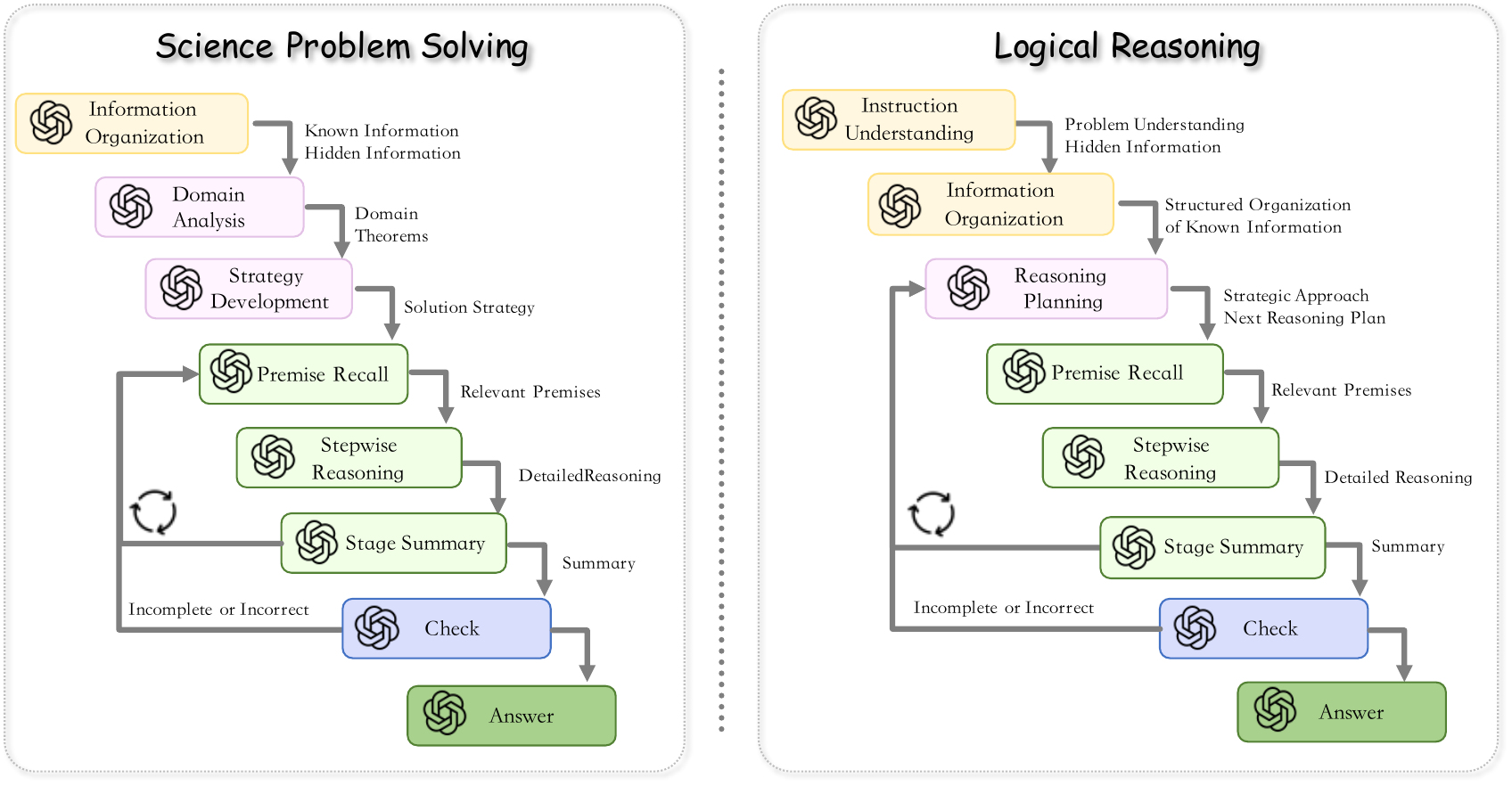}
  \includegraphics[width=2\columnwidth]{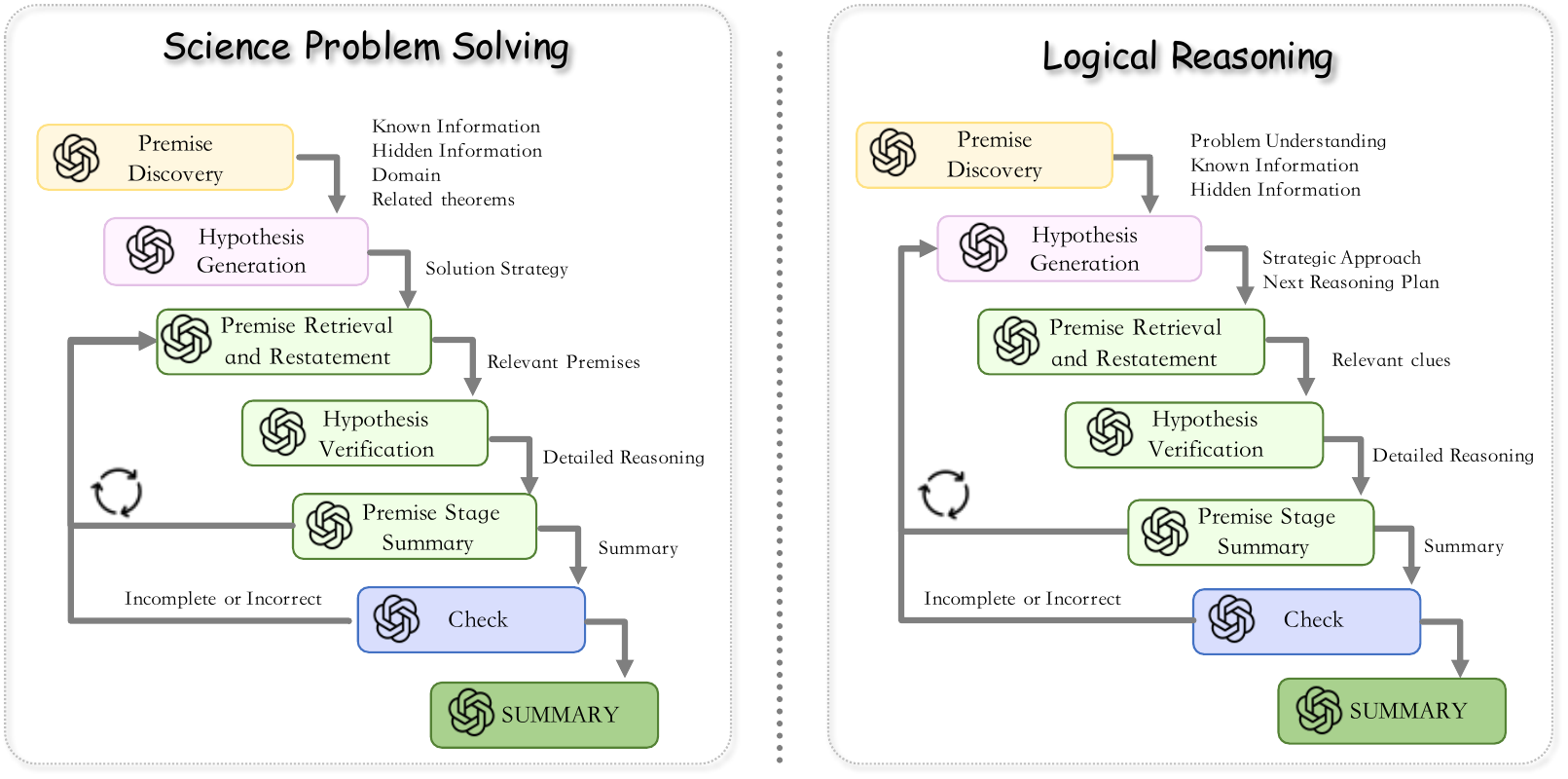}
  \caption{\textbf{Standard Operating Procedures (SOPs)} for Atomic Reasoning. The top subfigures show human cognitive patterns in Science Problem and Logical Reasoning tasks, while the bottom subfigures illustrate how these priors guide atomic reasoning actions for optimized execution.}
  \label{fig:sop}
\end{figure*}

SOP utilizes human expert prior knowledge in specific domains to optimize the execution of atomic reasoning. Figure~\ref{fig:sop} illustrates our SOP process: the top two subfigures depict common cognitive patterns and problem-solving strategies used by humans in Science Problem and Logical Reasoning tasks, while the bottom two subfigures illustrate how these domain-specific priors guide each action within the atomic reasoning framework. This enables systematic optimization in reasoning factor scheduling, ARA execution strategies, and example-based guidance, ultimately making the reasoning process more efficient and accurate.

\subsection{Context Compression in Reasoning Chains}
To mitigate the issue of redundant context during reasoning, we manage the Atomic Tree at the chain level, and compress the chain using LLMs once after a chain is completed, reducing cognitive overhead and improving the clarity of the reasoning process. This ensures that only the most relevant information is retained, facilitating more efficient reasoning without compromising accuracy.

\subsection{Design Details about Checker}

We introduce checker to enable more refined reflection. As shown in Figure~\ref{fig:checker}, we list the specific error types we have defined to guide the LLM, especially in conducting detailed checks for basic errors, thereby providing high-quality reasoning information for subsequent reasoning. In this section, we will introduce these error types in detail.

For premise-related step, the error types we define include:
\begin{itemize}
    \item \textbf{Content Conflict:} The error type refers to errors in the extracted premise information that directly conflict with the original information given in the question. To check for this, LLMs should compare the given premise information with the original question, ensuring that there are no discrepancies or contradictions between the two. If any mismatches are found, that would indicate a content conflict. 
    \item \textbf{Logical Contradiction:} The error involves logical contradictions between steps, where information may be inconsistent or fail to logically follow from previous steps due to misunderstandings. The LLMs checks for this by sequentially verifying each step to ensure logical consistency, comparing assumptions and conclusions with premises, and addressing any contradictions or gaps in reasoning, especially in conditional scenarios.
    \item \textbf{Expression Inconsistencies:} The error involves inconsistencies in the way expressions or equations are rewritten or understood across different steps. The LLMs is expected to check for such issues by comparing expressions between adjacent steps, verifying the accuracy of substitutions or transformations, and ensuring that symbols, units, and values remain consistent throughout the process.
\end{itemize}

For reasoning step, the error types we define include:
\begin{itemize}
    \item \textbf{Calculation Error:} The error involves miscalculations or transcription mistakes between consecutive equations. It suggests that the LLMs should verify each step of the calculation, check for arithmetic or functional errors, and ensure that intermediate results are correctly carried over between steps to avoid any discrepancies.
    \item \textbf{Common Sense Error:} The error involves violations of basic common sense, such as nonsensical or contradictory claims, like incorrect numerical comparisons or unrealistic assertions. The LLMs should identify these errors by checking conclusions and processes against fundamental common knowledge, such as ensuring numerical consistency.
    \item \textbf{Recapitulation Error:} The error occurs when an idea or concept is unnecessarily repeated or restated in a redundant manner. The LLMs can identify such errors by checking for repetitive statements within the reasoning process.
    \item \textbf{Ignoring of Premises:}
     The error involves neglecting specific constraints or overlooked scenarios from the premise, which can lead to incorrect conclusions. The LLMs is expected to detect such errors by ensuring that all constraints from the premise are checked.
    \item \textbf{Misusing of Premises:} 
    The error refers any conflicts or misinterpretations within a reasoning process where the statements deviate from the provided premise, either by confusing referential relationships or altering the premise information. The LLMs is expected to check this by directly comparing each statement to the premise for consistency.
    \item \textbf{Conclusion Errors:} The error concerns evaluating whether the conclusions in a reasoning process are logically derived from the previous steps and premises, and if they conflict with any given premises. The LLMs is expected to check this by mapping the conclusions back to the reasoning and premises, ensuring they are properly supported by evidence.
\end{itemize}
For ending step, the error types we define include:
\begin{itemize}
    \item \textbf{Result Omission:} The error occurs when a critical outcome or conclusion is left out in the final ending step, leading to incomplete or unclear results. The LLMs should identify this error by checking if all necessary conclusions or results are explicitly stated. 
    \item  \textbf{Results Inconsistency:} A results inconsistency error occurs when the outcome or conclusion is presented differently in various parts of the reasoning, creating contradictions. The LLMs should identify this error by comparing the results throughout the process.
    \item \textbf{Judgment Error:} The error occurs when an incorrect conclusion is suddenly reached in the final step, conflicting with the logic established earlier in the process. The LLMs should identify this error by ensuring that the final judgment aligns with the reasoning and conclusions drawn throughout the preceding steps, preventing any abrupt or illogical shifts.
    \item \textbf{Sorting Error:} The error refers to an inconsistency where the final output sequence deviates from the logical ordering established during intermediate reasoning steps. The LLMs should detect this discrepancy by cross-verifying the coherence between intermediate sorting logic and the final sequence output. 
\end{itemize}

\subsection{Prompt Design}
In the implementation of the Atomic Reasoner, there are two core roles played by the LLMs. The first role is the Routing Agent, which comprehensively understands and routes the reasoning process based on the current information in the Atomic Tree. The second role is the Reasoning Agent, which is tasked with detailed guidance on atomic reasoning actions provided by the routing agent. This agent performs thorough reasoning execution within the sub-solution domain.

Additionally, in the extra reflection mechanism, the Checker role plays a crucial part. The prompts for these LLM roles are meticulously defined to specify their roles and tasks. They also include necessary rule constraints and guidance tips, which we illustrate in the accompanying figures.

\section{Experiment Details}
\subsection{Baseline Algorithms Implementation}

Our implemented baseline algorithms include:

\begin{itemize}
    \item \textbf{CoT (Chain of Thought)}: Specifically, we implemented the naive zero-shot CoT. It is worth noting that in advanced models, the CoT capability (specifically short CoT) is inherently present in the model itself. Therefore, this result can be approximately considered as the model's inherent capability level.
    \item \textbf{SC-CoT (Self-Consistency Chain of Thought)}: We used 9 models and selected the final result based on their consensus voting.
    \item \textbf{ToT (Tree of Thought)}: We referred to the official implementation version \footnote{\url{https://github.com/princeton-nlp/tree-of-thought-llm}} and employ the BFS algorithm as the search strategy. The maximum search depth is set to $d = 8$, which controls the maximum number of steps for problem-solving, and the branching width is defined as $b = 5$, determining the breadth of the search. To validate the effectiveness of the branches, we perform $n_{eval} = 3$ parallel evaluations at each generation step. 
    The LLM evaluator outputs "yes", "likely", or "no", where "yes" is rewarded with $20$ points, "likely" with $1$ point, and "no" with no reward. The temperature parameter is set to $T=0.8$ to regulate the sampling strategy during the generation process. 
    \item \textbf{BoN (Best of N)}: This approach involves the LLM autonomously selecting the best result from multiple CoT attempts as the final result. The CoT attempts are generated with a temperature of $T_{gen} = 1.0$ to encourage diversity. Afterward, they are evaluated with a temperature of $T_{eval} = 0.0$ to ensure deterministic scoring. The process runs for $n = 6$, generating six candidates in total, from which the best answer is chosen based on the evaluation criteria.
    \item \textbf{G1}: Widely regarded as the first attempt to replicate o1. The method involves prompt engineering, where the large language model is prompted to self-reflect and propose multiple solutions to clone o1's behavior. We referred to the official implementation \footnote{\url{https://github.com/bklieger-groq/g1}}. We define the temperature as $T = 0.2$ and the maximum reasoning steps as $25$.
    \item \textbf{Llamaberry}: This method does not require training and combines the MCTS (Monte Carlo Tree Search) process. The official implementation can be found \footnote{\url{https://github.com/Jaimboh/Llamaberry-Chain-of-Thought-Reasoning-in-AI}}. We define the temperature as $T = 0.7$, which controls the level of randomness in the model's responses. We set the number of reasoning steps to $3$, indicating the number of reasoning steps the model will take.
    \item \textbf{Plan-MCTS}: This integrates MCTS into the planning stage. We referred to the official code from this project \footnote{\url{https://github.com/plan-mcts}}. The temperature is set to $T = 1.0$. The exploration weight is $1.0$ for the exploration-exploitation trade-off in UCT. The maximum tree depth is $d = 20$. The state score threshold is set to $0.8$, meaning states below this score are discarded. We perform $N_{rollout}$ = 8 rollouts. The LogicalConsistencyAgent, with a weight of $0.3$, scores the output based on its logical consistency, while the FeasibilityAgent, with a weight of $0.7$, evaluates the output's feasibility.
\end{itemize}

These baseline algorithms provide a comprehensive comparison for evaluating the performance and effectiveness of our Atomic Reasoner implementation.

\subsection{Dataset Introduction}

We validated our approach on 5 benchmarks, including:

\begin{enumerate}
    \item \textbf{AIME}: AIME is based on the high-difficulty mathematics competitions from the American Mathematics Competitions (AMC) series, transformed into a benchmark for evaluating advanced mathematical reasoning abilities of large language models. It tests the model's deep reasoning capabilities in algebra, geometry, number theory, etc.\footnote{\url{https://huggingface.co/datasets/AI-MO/aimo-validation-aime}}
    \item \textbf{MATH}~\cite{hendrycks2021math}: MATH is a benchmark focused on solving mathematical problems, including topics in algebra, calculus, probability, and more. It is divided into different difficulty levels to evaluate the model's mathematical logical reasoning and multi-step problem-solving abilities.
    \item \textbf{BBH (Big-Bench Hard)}~\cite{suzgun2022bbh}: BBH is a selection of 23 challenging tasks from the large-scale benchmark BigBench, covering areas such as logical reasoning and language understanding. It is used to test the model's generalization ability in complex scenarios, evaluating its performance in tasks like open-ended questions and counterfactual reasoning.
    \item \textbf{TheoremQA}~\cite{chen2023theoremqa}: TheoremQA is a benchmark focused on understanding and applying mathematical theorems. It requires the model to solve practical problems using theorem knowledge, covering areas such as geometry, algebra, and physics. It tests the model's mastery of mathematical theorems and their application in problem-solving.
    \item \textbf{ZebraGrid}~\cite{zebralogic2024}: ZebraGrid focuses on testing the language model's multi-step reasoning ability in \textbf{Logic Grid Puzzles}. This benchmark uses structured problem design, requiring the model to gradually eliminate contradictions based on multiple clues to deduce the only correct answer.
\end{enumerate}

Among these benchmarks, AIME and MATH primarily test the comprehensive mastery of fundamental mathematical knowledge and rigorous mathematical logical reasoning abilities. TheoremQA assesses a broader understanding and mastery of theorems. This means that the results on these benchmarks are largely constrained by the model's own knowledge system and the sufficiency of its knowledge base. Without a robust knowledge system, merely enhancing the reasoning process with prompts, without external knowledge or tools, will make it difficult to achieve effective performance improvements. BBH is a logical deduction problem with multiple-choice questions, while ZebraGrid involves fill-in-the-blank questions requiring the completion of a specified number of grid blanks. These two types of questions mainly assess natural language-based logical reasoning problems and do not require a specialized knowledge system.

\subsection{Generation of Data Subsets}

Considering the large scale of the test sets, which would consume significant computational resources (or API call costs) and time to complete multiple rounds of testing on all these benchmarks, we collected subsets for these tasks. The size of each collected subsets is approximately 100 questions. For most benchmarks, the process of collecting subsets was random. We believe that testing baseline algorithms and our methods on the same subsets is sufficient to fairly demonstrate the performance improvements. Moreover, this approach is widely adopted in academic research, such as \cite{slimLLM, genground}.

To lower the barriers and costs of academic research, we also encourage the community to adopt the same standard subsets for validation testing, thereby enhancing comparison efficiency. We have included some experimental data in the supplementary materials, which contains the results of all the questions in the subsets we collected. We will also make our experimental code public, which includes a unified interface for calling these benchmarks and their corresponding subsets.

\subsection{Training configuration}


We use LLamaFactory to fine-tune a Qwen2.5-1.5b-Instruct model, and adopt DeepSpeed ZeRO Stage 3 optimization to enable efficient full-parameter fine-tuning. The learning rate is set to $3e^{-6}$, with a cosine learning rate scheduler. Experiments are conducted on NVIDIA A100-PCIE-80GB GPUs, with a total training time of approximately 1.5 hours.

\section{More Discussion about Experiment Results}
\subsection{Fine-Tuning Results}
Based on the results of AR inference, we conducted data collection on the training set (distinct from the previously mentioned subset used for testing and evaluation), amassing a total of 1,000 small-scale data points. Due to computational resource constraints, we performed fine-tuning tests exclusively on a smaller-scale language model, Qwen2.5-1.5b-Instruct. The results indicated a significant performance improvement compared to the base model.

It is important to note that although the table includes the results of R1 fine-tuning on the 1.5b model, this comparison is not entirely rigorous and does not imply that our model outperforms R1. Achieving such a goal is challenging at this stage but remains a worthwhile objective. According to~\citep{guo2025deepseekr1}, R1 is distilled based on the Qwen2.5-1.5b-Math model. We also attempted to test and fine-tune this model, finding that its performance in non-mathematical tasks was significantly inferior to Qwen2.5-1.5b-Instruct. Our attempts at fine-tuning revealed that this model's 4k context limit imposed higher requirements on our data cleaning process, making it difficult to distill the slow-thinking process into this mathematical model. Consequently, we did not pursue further attempts with the math model.

That table aims to illustrate another phenomenon: fine-tuning on the same 1.5b scale model can elevate performance to a level comparable to or even surpassing the R1 distillation model of the same scale. This underscores the potential of the data generated by our algorithm. We also believe that with further data cleaning, the performance gains from fine-tuning could be even greater. However, this diverges somewhat from our initial research objective. We aimed to demonstrate—and we believe we have sufficiently proven—the capability and potential of AR in inference scaling.

\clearpage

\begin{figure*}
\begin{tcolorbox}[colback=gray!10, colframe=black, width=\textwidth, arc=2mm, boxrule=0.5mm, left=2mm, right=2mm, top=5mm, bottom=2mm, 
title=\textbf{Prompt for the Routing Agent - (Expansion)}
]

You are an expert tasked with guiding a solver through a structured reasoning process to solve a complex problem. Solving this problem requires deeply and thoroughly extracting information from the problem statement and reasoning process, exploring all plausible approaches systematically and verifying them with rigorous logical precision. This complex process requires you to provide step-by-step guidance.
\\

Specifically, you will be presented with the problem statement and the complete reasoning steps up to now; you need to analyze what reasoning action should be executed next (such as proposing a hypothesis or conducting a self-check), and provide detailed methodological guidance to instruct the **solver** on how to execute this action specifically, thereby completing the task step by step.
\\

\# The reasoning factor actions you can define for the next step are as follows:

\textbf{\#\# First Category: Premise-related Reasoning Factors, aimed at enhancing information collection ability and attention to information}

A premise refers to the known conditions, assumptions, or facts that serve as the foundation or starting point in the reasoning or argumentation process. Premises provide the necessary background information and constraints for the reasoning process and are the basis for reasoning and drawing conclusions.

\textbf{1. **Premise Discovery**}

    - This is the starting point of the whole reasoning process. The solver needs to **identify the basic conditions and constraints of the problem or task, extract necessary rules or constraints, and organize implicit rule constraints**. This step helps the solver clarify the background and scope of the problem.

    - Premise discovery clarifies the foundation of the reasoning process and prevents incorrect assumptions.

...
\\

\textbf{\#\# Second Category: Slow Thinking Reasoning Factors, aimed at guiding detailed thinking processes}

A hypothesis is a possible explanation, solution, or assumption proposed based on premises. Hypotheses are usually preliminary answers to unknown problems, not yet verified, and require further reasoning and evidence support.

\textbf{1. **Hypothesis Generation**}

   - You need to carefully examine the existing problem-solving steps and the utilization of existing premises, then guide the solver to focus on unresolved sub-steps, relevant premises, or ambiguities in the problem statement, and then propose necessary possible hypotheses for a specific sub-problem or blank.

   - Your role is to guide the solver in proposing hypotheses, **rather than directly providing hypotheses yourself**.

...
\\

\textbf{\#\# Third Category: Ending Reasoning Action **SUMMARY<FINISHED>** }(Cautiously! Do not choose this action unless necessary!)

- [Attention] **Mandatorily**, The first time you believe the reasoning has successfully concluded, you should force the solver to perform **Hypothesis Verification** action until it is truly confirmed as successfully completed.

- If all the substeps have been completed with no errors found in the process and results upon verification, then please choose this ending action **SUMMARY<FINISHED>** at the current moment.

\newpage
\end{tcolorbox}
\end{figure*}

\begin{figure*}
\begin{tcolorbox}[colback=gray!10, colframe=black, width=\textwidth, arc=2mm, boxrule=0.5mm, left=2mm, right=2mm, top=5mm, bottom=2mm, 
title=\textbf{Prompt for the Routing Agent - (Expansion) - (Continued)}
]

\# You need to flexibly schedule these actions and ultimately achieve task resolution. Here are some rules about scheduling these actions:

1. When the current reasoning process is complex, making it difficult to clearly analyze the current state and propose hypotheses, conducting premise retrieval and restatement, premise summaries, etc., is very necessary. This can guide the solver to sort out clues and enhance the emphasis on relevant information.

...
\\

\# The problem that needs to be solved is:

<Problem Description>

\end{tcolorbox}
\end{figure*}

\begin{figure*}
\begin{tcolorbox}[colback=gray!10, colframe=black, width=\textwidth, arc=2mm, boxrule=0.5mm, left=2mm, right=2mm, top=5mm, bottom=2mm, 
title=\textbf{Prompt for the Reasoning Agent}
]

You are a solver engaged in a dialogue with an expert, solving a complex problem under the expert's guidance. Solving this problem requires you to deeply and thoroughly extract information from the problem statement and reasoning process, explore as many approaches as possible, and verify them as precisely as possible. This complex process requires you to follow the expert's step-by-step guidance to achieve detailed and accurate solutions.
\\

Specifically, you will be presented with the problem statement, the complete reasoning steps up to the current point, and the expert's guidance for the current step. Based on this, you need to execute the reasoning actions instructed by the expert, accurately and comprehensively completing the solution for the current step.
\\

You must note:

1. Each step of your reasoning process should be detailed enough to avoid missing any critical information.

2. You do not need to perform additional actions; always carefully implement and ensure the correctness of the reasoning process as much as possible. Always keep in mind the relevant rules, implicit or explicit constraints, and formatting restrictions related to problem-solving.

3. Ensure each clue is fully understood and its implications are explicitly identified before generating hypotheses.

4. Mark any unverified information as pending verification or assumption to avoid it being mistaken for a correct premise.

5. All statements that have not been verified correctly must be explicitly marked as assumptions! It is not allowed to use definitive or certain language to express your assumptions during the hypothesis generation phase.
\\

\# The problem that needs to be solved is:

<Problem Description>

\end{tcolorbox}
\end{figure*}

\begin{figure*}
\begin{tcolorbox}[colback=gray!10, colframe=black, width=\textwidth, arc=2mm, boxrule=0.5mm, left=2mm, right=2mm, top=5mm, bottom=2mm, 
title=\textbf{Prompt for the Routing Agent - (Backtracing)}
]

You are a character for routing, specializing in examining the entire process of complex logical reasoning from a macro perspective. At the conclusion of one reasoning chain, you provide suggestions to guide a return to a specific node on the historical reasoning path, initiating new thinking to better solve the problem.
\\

You will be informed of the complex problem to be solved and the entire historical reasoning process (presented in a tree structure). You need to review this process and then provide suggestions on which node in the current chain should be revisited to start a new reasoning branch.
\\

Your process of reviewing and providing suggestions can follow these guidelines:

1. The situations you encounter fall into two categories:

    - The current reasoning chain has been completed, and you need to guide a return to a node to start a new branch, exploring better results.
    
    - The current reasoning chain has reached a dead end, making it difficult to continue, requiring a return to a node to start a new branch, avoiding the dead end, and exploring better results.
    \\
    
2. The target node to which you need to return can generally be considered in three categories:

    - Return to intermediate nodes related to incorrect content: You may find that the results of certain reasoning stages are incorrect, or some intermediate conclusions do not match reality, so you can return to the node that produced this intermediate conclusion to start a more correct branch.
    
    - Return to key intermediate nodes affecting the current reasoning result: Some nodes on the current reasoning path may have better actions to achieve better results.
    
    - Intermediate nodes with unexplored branches: Some nodes on the current reasoning path may have branches that have not yet been activated.
\\

Currently, the existing reasoning process and conclusions are summarized as follows, presented in a tree structure:
\\

<existing process>
\\

Among them, you need to backtrack to a node on the current chain, which is:
\\

<current chain>
\end{tcolorbox}
\end{figure*}

\begin{figure*}
\begin{tcolorbox}[colback=gray!10, colframe=black, width=\textwidth, arc=2mm, boxrule=0.5mm, left=2mm, right=2mm, top=5mm, bottom=2mm, 
title=\textbf{Prompt for the Checker Agent}
]

Please review the following multi-step reasoning process thoroughly. Focus on six potential error types and analyze the process in detail. For each type, explicitly identify errors or confirm correctness with detailed reasoning. When checking for each type of error, you can use the recommended checking methods for the corresponding error. If errors are present, explain their nature and suggest corrections.
\\

\textbf{**Defined error types:**}

\textbf{1. **Logical Contradictions**:}

   Check for logical contradictions between steps. Ensure the information in each step is consistent and follows logically from the previous one. Look for errors caused by misunderstandings or misinterpretations of prior steps.
   \\
   
\textbf{**Suggested Checking Methods**:}
   
- Break down the multi-step reasoning process and verify each step sequentially to ensure it logically follows from the previous step. 
   
- Compare the reasoning, assumptions, and conclusions of each step with its premises to ensure they are consistent.
   
- For conditional branches or hypotheses, check each possibility individually to confirm there are no overlooked or contradictory assumptions. 
   
- Identify obvious causal errors or gaps in reasoning to ensure a coherent logical chain.

...
\\

For each error type, provide a detailed analysis and justify your findings, whether errors are present or not. If errors are identified, describe them clearly and suggest solutions. Ensure every aspect of the process is reviewed comprehensively. In the summary, you should give the check result and suggestions for modifications if there is error.
\\

Process to be examined:

<reasoning process>

\end{tcolorbox}
\end{figure*}
\begin{figure*}
\begin{tcolorbox}[colback=gray!10, colframe=black, width=\textwidth, arc=2mm, boxrule=0.5mm, left=2mm, right=2mm, top=5mm, bottom=2mm, 
title=\textbf{Case 1: AR-driven Slow-thinking Reasoning in Task Logical Deduction of BBH (1/3)}
]
\subsection*{Task}
Solve the logical deduction task which requires deducing the order of a sequence of objects:\\
The following paragraphs each describe a set of seven objects arranged in a fixed order. The statements are logically consistent within each paragraph. On a branch, there are seven birds: a hummingbird, a cardinal, a blue jay, an owl, a raven, a quail, and a robin. The hummingbird is to the left of the quail. The robin is to the left of the cardinal. The blue jay is the leftmost. The cardinal is the fourth from the left. The raven is the third from the right. The owl is the third from the left.

Options:

(A) The hummingbird is the second from the right

(B) The cardinal is the second from the right

(C) The blue jay is the second from the right\\
(D) The owl is the second from the right

(E) The raven is the second from the right

(F) The quail is the second from the right

(G) The robin is the second from the right

Your final answer should follow this format: "The correct answer is (insert answer here)".

\noindent\rule{\textwidth}{0.5pt}
\subsection*{Reasoning Output directly by GPT-4o-mini}

Let's go through the information step by step:

1. **The blue jay is the leftmost**: So, the blue jay is in the first position.

2. **The owl is the third from the left**: This places the owl in the third position.

3. **The cardinal is the fourth from the left**: So, the cardinal is in the fourth position.

4. **The raven is the third from the right**: This places the raven in the fifth position.

5. **The hummingbird is to the left of the quail**: The hummingbird must be in a position to the left of the quail.

6. **The robin is to the left of the cardinal**: The robin must be in a position to the left of the cardinal.

Given this information, we can deduce the following order:

1. **Blue jay** (leftmost, first position)

2. **Hummingbird** (must be placed before the quail and to the left of the robin)

3. **Owl** (third position)

4. **Cardinal** (fourth position)

5. **Raven** (fifth position)

6. **Quail** (to the right of the hummingbird, but not the rightmost)

7. **Robin** (must be the second from the right)

Now, we can check the options:

- The **robin** is the second from the right, so the correct answer is:
  
**The correct answer is (G)**.

\paragraph{Evaluation Result}

\textbf{Incorrect}

\end{tcolorbox}
\end{figure*}

\begin{figure*}
\begin{tcolorbox}[colback=gray!10, colframe=black, width=\textwidth, arc=2mm, boxrule=0.5mm, left=2mm, right=2mm, top=5mm, bottom=2mm, 
title=\textbf{Case 1: AR-driven Slow-thinking Reasoning in Task Logical Deduction of BBH (2/3)}
]

\subsection*{Reasoning Output following AR by GPT-4o-mini}
\paragraph{Step 1: Premise Discovery}

The problem requires us to determine the order of seven birds based on specific positional clues.

From the problem statement, we extract the following clue information:

- **Clue 1:** Hummingbird (H) is to the left of Quail (Q).  

- **Clue 2:** Robin (R) is to the left of Cardinal (C). 

- **Clue 3:** Blue Jay (BJ) is the leftmost (position 1).

- **Clue 4:** Cardinal (C) is the fourth from the left (position 4). 

- **Clue 5:** Raven (Rav) is the third from the right (position 5). 

- **Clue 6:** Owl (O) is the third from the left (position 3).

Each clue establishes a clear positional framework that will guide the arrangement of the birds.

\paragraph{Step 2: Hypothesis Generation}

Propose hypotheses, considering possible solutions, derivations, and approaches for the problem.

- **Hypothesis 1:** The arrangement of the birds could be:  
     1. Blue Jay (BJ)  
     2. Hummingbird (H)  
     3. Owl (O)  
     4. Cardinal (C)  
     5. Raven (Rav)  
     6. Quail (Q)  
     7. Robin (R)  
     This arrangement satisfies all clues, with H to the left of Q and R to the left of C.

   - **Hypothesis 2:** The arrangement of the birds could be:  
     1. Blue Jay (BJ)  
     2. Robin (R)  
     3. Owl (O)  
     4. Cardinal (C)  
     5. Raven (Rav)  
     6. Hummingbird (H)  
     7. Quail (Q)  
     This arrangement also satisfies all clues, with R to the left of C and H to the left of Q.

\paragraph{Step 3: Hypothesis Verification}

We will check Hypothesis 1, which proposes the following order: Blue Jay (BJ), Hummingbird (H), Owl (O), Cardinal (C), Raven (Rav), Quail (Q), and Robin (R).

Upon verification, Clue 2 is not satisfied since the robin is not to the left of the cardinal.

Consequently, Hypothesis 1 is rejected, prompting a reassessment of the alternative hypotheses for further verification.

The problem requires us to verify the arrangement of seven birds based on specific clues.

\paragraph{Step 4: Hypothesis Verification}

We will check Hypothesis 2, which proposes the following order: Blue Jay (BJ), Robin (R), Owl (O), Cardinal (C), Raven (Rav), Hummingbird (H), and Quail (Q).

Upon verification, all clues are satisfied: the hummingbird is to the left of the quail, the robin is to the left of the cardinal, the blue jay is the leftmost, the cardinal is fourth from the left, the raven is third from the right, and the owl is third from the left.

Thus, the verification process confirms that Hypothesis 2 is correct and meets all the conditions set forth in the problem statement.

The final arrangement confirmed that the raven is the second from the right.

The correct answer is (E) The raven is the second from the right.

\end{tcolorbox}
\end{figure*}

\begin{figure*}
\begin{tcolorbox}[colback=gray!10, colframe=black, width=\textwidth, arc=2mm, boxrule=0.5mm, left=2mm, right=2mm, top=5mm, bottom=2mm, 
title=\textbf{Case 1: AR-driven Slow-thinking Reasoning in Task Logical Deduction of BBH (3/3)}
]
\subsection*{Reasoning Output following AR by GPT-4o-mini}
\paragraph{Step 4: CHECKER}

To verify the reasoning process regarding the identification of the second from the right bird in the final arrangement, we will follow the outlined steps:

1. **Re-sort the sequence as needed:**  

Since we are looking for the "second from the right," we will sort the sequence from right to left (largest to smallest).

2. **Explicitly list the re-sorted sequence:**  

The final arrangement of the birds is:  
   1. Blue Jay (BJ)  
   2. Robin (R)  
   3. Owl (O)  
   4. Cardinal (C)  
   5. Raven (Rav)  
   6. Hummingbird (H)  
   7. Quail (Q)

When sorted from right to left, the sequence becomes:  
   1. Quail (Q)  
   2. Hummingbird (H)  
   3. Raven (Rav)  
   4. Cardinal (C)  
   5. Owl (O)  
   6. Robin (R)  
   7. Blue Jay (BJ)  

3. **Define the meaning of positional terms clearly:**  

"Second from the right" refers to the item ranked second in the sequence re-sorted from right to left.

4. **Verify step-by-step whether the designated item matches the description of its position in the re-sorted sequence:**  

In the re-sorted sequence: 

   - 1st from the right: Quail (Q)  
   
   - 2nd from the right: Hummingbird (H) 
   
5. **Cross-check the reasoning to ensure the stated item aligns with its actual position based on the re-sorted sequence:** 

The reasoning states that the second from the right is the Raven (Rav), but in the re-sorted sequence, the second from the right is actually the Hummingbird (H).

**Check Result: There is an error.**

The correct answer is that the second from the right is the Hummingbird (H), not the Raven (Rav).

\paragraph{Step 4: REVISED CONTENT}

We will check Hypothesis 2, which proposes the following order: Blue Jay (BJ), Robin (R), Owl (O), Cardinal (C), Raven (Rav), Hummingbird (H), and Quail (Q).

Upon verification, all clues are satisfied: the hummingbird is to the left of the quail, the robin is to the left of the cardinal, the blue jay is the leftmost, the cardinal is fourth from the left, the raven is third from the right, and the owl is third from the left.

Thus, the verification process confirms that Hypothesis 2 is correct and meets all the conditions set forth in the problem statement.

However, upon further review of the final arrangement, the second from the right is not the Raven (Rav), but the Hummingbird (H).

The correct answer is (A) The Hummingbird is the second from the right, not the Raven.

\paragraph{Final Conclusion}
The correct answer is (A) The hummingbird is the second from the right.

\paragraph{Evaluation Result}
\textbf{Correct}

\end{tcolorbox}
\end{figure*}

\begin{figure*}
\begin{tcolorbox}[colback=gray!10, colframe=black, width=\textwidth, arc=2mm, boxrule=0.5mm, left=2mm, right=2mm, top=5mm, bottom=2mm, 
title=\textbf{Case 2: AR-driven Slow-thinking Reasoning in Task of ZebraGrid (1/4)}
]
\subsection*{Task}
There are 3 houses, numbered 1 to 3 from left to right, as seen from across the street. Each house is occupied by a different person. Each house has a unique attribute for each of the following characteristics:

- Each person has a unique name: `Arnold`, `Eric`, `Peter`
 
- People have unique favorite book genres: `romance`, `mystery`, `science fiction`
 
- Everyone has something unique for lunch: `grilled cheese`, `pizza`, `spaghetti`
 
- Everyone has a favorite smoothie: `desert`, `watermelon`, `cherry`

\vspace{0.35cm}
\textbf{\#\# Clues:}

1. The person who loves mystery books is somewhere to the left of the person who likes Cherry smoothies.

2. Eric is in the second house.

3. Peter is in the first house.

4. Eric is the person who loves romance books.

5. The Desert smoothie lover is in the third house.

6. The person who is a pizza lover is Arnold.

7. The person who loves the spaghetti eater and the person who likes Cherry smoothies are next to each other.

\noindent\rule{\textwidth}{0.5pt}
\subsection*{Reasoning Output directly by GPT-4o-mini}

Based on the clues, let’s analyze the puzzle step by step:

1. **Clue 2**: Eric is in House 2:
   - House 1: ?
   - House 2: Eric
   - House 3: ?

2. **Clue 3**: Peter is in House 1:
   - House 1: Peter
   - House 2: Eric
   - House 3: ?

3. **Clue 4**: Eric loves romance books:
   - House 1: Peter
   - House 2: Eric (Romance)
   - House 3: ?

4. **Clue 6**: Arnold loves pizza, so he’s in House 3:
   - House 1: Peter
   - House 2: Eric (Romance)
   - House 3: Arnold (Pizza)

5. **Clue 5**: The Desert smoothie lover is in House 3, so Arnold loves it:
   - House 1: Peter
   - House 2: Eric (Romance)
   - House 3: Arnold (Pizza, Desert)

6. **Clue 1**: Peter loves mystery books and Eric loves Cherry smoothies:
   - House 1: Peter (Mystery)
   - House 2: Eric (Romance, Cherry)
   - House 3: Arnold (Pizza, Desert)

7. **Clue 7**: Arnold loves spaghetti, so:
   - House 1: Peter (Mystery)
   - House 2: Eric (Romance, Cherry)
   - House 3: Arnold (Pizza, Desert, Spaghetti)

8. Arnold loves pizza and spaghetti, so Peter must love grilled cheese.

Final answer:

- House 1: Peter (Mystery, Grilled Cheese)

- House 2: Eric (Romance, Cherry)

- House 3: Arnold (Pizza, Desert, Spaghetti)

\paragraph{Evaluation Result}
75\%

\end{tcolorbox}
\end{figure*}

\begin{figure*}
\begin{tcolorbox}[colback=gray!10, colframe=black, width=\textwidth, arc=2mm, boxrule=0.5mm, left=2mm, right=2mm, top=5mm, bottom=2mm, 
title=\textbf{Case 2: AR-driven Slow-thinking Reasoning in Task of ZebraGrid (2/4)}
]
\subsection*{Reasoning Output following AR by GPT-4o-mini}

\paragraph{Step 1: Premise Discovery}
The problem involves determining the attributes of three individuals, Arnold, Eric, and Peter, based on specific clues. The attributes to be identified include their names, favorite book genres, lunch preferences, and favorite smoothies. The explicit conditions documented include: 

1. The mystery book lover is to the left of the cherry smoothie lover. 

2. Eric is confirmed to be in the second house. 

3. Peter is confirmed to be in the first house. 

4. Eric loves romance books. 

5. The desert smoothie lover resides in the third house. 

6. Arnold is identified as the pizza lover. 

7. The spaghetti lover is next to the cherry smoothie lover. 

The clues are classified into categories: location clues, book genre clues, lunch preferences, and smoothie preferences, which aids in organizing the information and understanding the relationships between the attributes of the individuals.

\paragraph{Step 2: Hypothesis Generation}

1. Analyze the premises of the current sub-step.

Existing reasoning:

- Peter is in House 1.

- Eric is in House 2 and loves romance books.

- The desert smoothie lover is in House 3.

There are also clues to unused information, including:

- Clue 1: The mystery book lover is somewhere to the left of the cherry smoothie lover.

- Clue 5: The spaghetti lover and the cherry smoothie lover are next to each other.

- Clue 6: Arnold is the pizza lover.

2. Propose hypotheses.

- Hypothesis 1: Next, we can determine the favorite book genres of the remaining individuals, particularly focusing on the mystery book lover and how they relate to the cherry smoothie lover.

- Hypothesis 2: We can determine the food preferences next, especially since we know Arnold is the pizza lover, which may help us identify the remaining food preferences.

We choose Hypothesis 1 to focus on determining the favorite book genres of each person, particularly the mystery book lover and their relationship to the cherry smoothie lover. This is because the clues about the book genres directly relate to the positions of the individuals and can help clarify the remaining attributes.

\end{tcolorbox}
\end{figure*}

\begin{figure*}
\begin{tcolorbox}[colback=gray!10, colframe=black, width=\textwidth, arc=2mm, boxrule=0.5mm, left=2mm, right=2mm, top=5mm, bottom=2mm, 
title=\textbf{Case 2: AR-driven Slow-thinking Reasoning in Task of ZebraGrid (3/4)}
]
\subsection*{Reasoning Output following AR by GPT-4o-mini}

\paragraph{Step 3: Hypothesis Verification} 

1. Sequentially Check All Known Clues Against the Current Hypothesis

**Clue 1**: The mystery book lover is to the left of the cherry smoothie lover.

- Since Eric is in House 2 and loves romance books (Clue 4), he cannot be the mystery book lover. Therefore, the mystery book lover must be either Peter (House 1) or Arnold (House 3).

- If Peter is the mystery book lover, then the cherry smoothie lover must be in House 2 or 3. However, since Eric loves romance, he cannot be the cherry smoothie lover. Thus, Arnold must be the cherry smoothie lover, which contradicts Clue 6 (Arnold is the pizza lover).

- Therefore, Peter must be the mystery book lover, and Arnold must be the cherry smoothie lover.

**Clue 5**: The Desert smoothie lover is in the third house.

- This means Arnold cannot be the desert smoothie lover, confirming he is the cherry smoothie lover.

**Clue 6**: The person who is a pizza lover is Arnold.

- This confirms that Arnold is not the mystery book lover, as he is the pizza lover.

**Clue 7**: The person who loves spaghetti and the person who likes cherry smoothies are next to each other.
   
- Since Arnold is the cherry smoothie lover, the spaghetti lover must be in House 2 (Eric) or House 1 (Peter). However, since Eric loves romance books, he cannot be the spaghetti lover. Therefore, Peter must be the spaghetti lover.

2. Summary of Findings

From the analysis:

- Peter is the mystery book lover and lives in House 1.

- Eric, who loves romance books, lives in House 2.

- Arnold, who is the pizza lover, lives in House 3 and is the cherry smoothie lover.

- The desert smoothie lover must be Eric, as he is the only one left.

Thus, we have verified the relationships between the individuals and their attributes based on the clues provided.

\paragraph{Step 4: SUMMARY<FINISHED>}

Reasoning:
Based on the clues, we determined that Peter is in House 1 and loves mystery books. Eric is in House 2 and loves romance books, while Arnold is in House 3 and loves pizza. The desert smoothie lover is in House 3, which means Arnold loves the desert smoothie. Since the cherry smoothie lover must be next to the spaghetti lover, and since Arnold is the cherry smoothie lover, Peter must be the one who loves spaghetti. Therefore, Eric must love watermelon smoothies.

Solution:

- House 1: Peter (mystery, spaghetti, watermelon)

- House 2: Eric (romance, grilled cheese, watermelon)

- House 3: Arnold (science fiction, pizza, desert)

\paragraph{[Backward to]Step 2: Hypothesis Generation}

ANALYSIS: After reviewing the current reasoning chain, I noticed that node 2 is an intermediate node with unexplored branches. At this node, we confirmed certain attributes based on the clues, but we did not fully explore the implications of the adjacency clues regarding the cherry smoothie lover and the spaghetti lover. The assumptions made in this node led to a conclusion that Arnold is the cherry smoothie lover, which contradicts the established clues. Backtracking to this node allows us to reassess the relationships and explore alternative configurations that could lead to a more accurate solution.

\end{tcolorbox}
\end{figure*}

\begin{figure*}
\begin{tcolorbox}[colback=gray!10, colframe=black, width=\textwidth, arc=2mm, boxrule=0.5mm, left=2mm, right=2mm, top=5mm, bottom=2mm, 
title=\textbf{Case 2: AR-driven Slow-thinking Reasoning in Task of ZebraGrid (4/4)}
]
\subsection*{Reasoning Output following AR by GPT-4o-mini}

\paragraph{Step 2: Hypothesis Generation}
1. **Analyze the premises of the current sub-step**.

- From the clues, we know that Peter is in House 1, Eric is in House 2, and Arnold is in House 3. 

- Eric loves romance books, and Arnold is the pizza lover.

- The desert smoothie lover is in House 3, which is Arnold.

- The mystery book lover must be to the left of the cherry smoothie lover (Clue 1), and the spaghetti lover and cherry smoothie lover are next to each other (Clue 7).

- Given that Peter is in House 1, he must be the mystery book lover, which places the cherry smoothie lover in House 2 (Eric).

- This means that the spaghetti lover must be in House 3 (Arnold).

2. **Propose hypotheses**.

- **Hypothesis 1**: Peter loves mystery books, Eric loves romance books, and Arnold loves science fiction.

- **Hypothesis 2**: Peter loves spaghetti, Eric loves grilled cheese, and Arnold loves pizza.

- **Hypothesis 3**: Peter loves cherry smoothies, Eric loves watermelon smoothies, and Arnold loves desert smoothies.

We'll proceed to verify these before moving to the next step.

\paragraph{Step 3: Hypothesis Verification}

1. Sequentially Check All Known Clues Against the Current Hypothesis

- **Clue 1**: The mystery book lover (Peter) is to the left of the cherry smoothie lover (Eric). This is consistent as Peter is in House 1 and Eric is in House 2.

- **Clue 2**: Eric is in the second house. This is confirmed.

- **Clue 3**: Peter is in the first house. This is confirmed.

- **Clue 4**: Eric loves romance books. This is confirmed.

- **Clue 5**: The desert smoothie lover is in the third house. This is consistent with Arnold being in House 3.

- **Clue 6**: The person who is a pizza lover is Arnold. This is confirmed as Arnold is in House 3.

- **Clue 7**: The spaghetti lover (Peter) and the cherry smoothie lover (Eric) are next to each other. This is consistent as Peter is in House 1 and Eric is in House 2.

2. Summary of Findings

All hypotheses regarding the favorite book genres, food preferences, and smoothie preferences have been verified against the clues and are consistent. Peter is the mystery book lover, Eric loves romance books, and Arnold is the pizza lover. The smoothie preferences align as well, with Arnold loving the desert smoothie, and Eric loving cherry smoothies. Therefore, the current hypotheses are valid and consistent with all provided clues.

\paragraph{Step 4: SUMMARY<FINISHED>}

Reasoning: The final arrangement has Peter in House 1, where he loves spaghetti and is the mystery book lover, Eric in House 2, who enjoys romance books and drinks cherry smoothies, and Arnold in House 3, who loves desert smoothies and is the pizza lover.

Solution:

- House 1: Peter (mystery, spaghetti, watermelon)

- House 2: Eric (romance, grilled cheese, cherry)

- House 3: Arnold (science fiction, pizza, desert)

\paragraph{Final Conclusion}
Solution:

- House 1: Peter (mystery, spaghetti, watermelon)

- House 2: Eric (romance, grilled cheese, cherry)

- House 3: Arnold (science fiction, pizza, desert)

\paragraph{Evaluation Result}
\textbf{100}

\end{tcolorbox}
\end{figure*}

\clearpage
\end{document}